\documentclass[12pt,3p]{elsarticle}
\usepackage[utf8]{inputenc}

\usepackage[namelimits,sumlimits]{amsmath}
\usepackage{amsfonts,amssymb,amsthm}
\usepackage{cancel} 
\usepackage{soul} 
\usepackage{graphicx}
\graphicspath{{images/}}
\usepackage[hidelinks]{hyperref}
\usepackage{fancyhdr}
\usepackage{url}
\usepackage{array}
\usepackage{dirtree}
\usepackage{enumitem}
\usepackage[explicit]{titlesec}
\usepackage{etoolbox}
\usepackage{acro}
\usepackage{xcolor}
\usepackage{subcaption}
\newsavebox{\imagebox}
\usepackage{blindtext}
\usepackage{csquotes}
\usepackage{pgfgantt}
\usepackage[export]{adjustbox}
\usepackage{ragged2e}
\usepackage{pdflscape} 
\usepackage{tikz}
\usetikzlibrary{shapes.geometric, arrows}
\usetikzlibrary{arrows.meta}
\usetikzlibrary{quotes,angles}
\usepackage{textpos}
\usepackage[linesnumbered,ruled,vlined]{algorithm2e}
\usepackage{float}
\usepackage{import}
\usepackage{algorithmicx}

\newcommand{\captioncomment}[2]{\caption[{#1}]{#1. #2 }}

\newcommand{\abbrev}[2]{\DeclareAcronym{#2}{
  short = #2,
  long  = #1,
  class = abbreviations
}}
\newcommand{\notat}[3]{\DeclareAcronym{#2}{
  short = #3,
  long  = #1,
  class = notations
}}

\newcommand{\norm}[1]{\left\lVert#1\right\rVert}
\newcommand{\point}[1]{(\textit{#1})}
\newcommand{\changes}[1]{\textcolor{black}{#1}}
\newenvironment{rcases}
  {\left.\begin{aligned}}
  {\end{aligned}\right\rbrace}

\abbrev{degrees of freedom}{dof}
\abbrev{inverse geometric solution}{IGS}
\abbrev{direct geometric solution}{DGS}
\abbrev{Denavit Hartenberg}{DH}
\abbrev{centre of rotation}{COR}
\abbrev{end-effector}{EE}
\abbrev{collision free workspace}{CFW}
\abbrev{forward instantaneous kinematic problem}{FIKP}
\abbrev{inverse instantaneous kinematic problem}{IIKP}
\abbrev{Universal joint}{U}
\abbrev{Spherical joint}{S}
\abbrev{Spherical Parallel Mechanism}{SPM}
\abbrev{Rotational joint}{R}
\abbrev{Genetic Algorithm}{GA}
\abbrev{Nelder-Mead}{NM}
\abbrev{Operating Room}{OR}
\abbrev{Centre Hospitalier Universitaire}{CHU}
\abbrev{Parallel Kinematic Manipulator}{PKM}
\abbrev{Non-dominated Sorting Genetic Algorithm}{NSGA-II}
\abbrev{Multi-Objective Evolutionary Algorithm based on Decomposition}{MOEA/D}
\abbrev{Particle Swarm Optimization}{PSO}
\abbrev{Multi Objective Optimization}{MOO}
\abbrev{Differential Evolution}{DE}
\abbrev{Regular Dextrous Workspace}{RDW}
\abbrev{Spherical-Spherical}{SS}
\abbrev{Universal-Spherical}{US}
\abbrev{Revolute-Revolute}{RR}

\DeclareAcronym{CAD_model}{
  short = CAD modeling,
  long  = Computer Aided Design,
  class = abbreviations
}
\notat{generalized output vector}{op_vec}{$\mathbf{x}$}
\notat{input vector}{ip_vec}{$\mathbf{q_i}$}
\notat{slack/surplus variable vector}{var_vec}{$\mathbf{s}$}
\notat{configuration space}{c_space}{$\mathcal{D}$}
\notat{input space}{ip_space}{$\mathcal{I}$}
\notat{output space}{op_space}{$\mathcal{K}$}
\notat{optimisation space}{o_space}{$\mathcal{O}$}
\notat{special Euclidean group of dimension 3}{se3}{\textit{SE}(3)}
\notat{special orthogonal group of dimension 3}{so3}{\textit{SO}(3)}
\notat{pose of the mechanism}{pose}{$\mathbf{x}$}
\notat{actuation joint vector}{act_q}{$\mathbf{q}_a$}
\notat{output twists}{op_twist}{$\mathbf{t}$}
\notat{feasible set}{FS}{$\mathcal{F}$}
\notat{quality index}{qi}{$\kappa^{-1}$}
\notat{manipulability ellipsoid volume}{ev}{$v_e$}
\notat{conditioning number}{cnum}{$\kappa$}
\notat{Jacobian matrix}{jac}{$\mathbf{J}$}
\notat{global conditioning index}{gqi}{$\kappa^{-1}_{g}$}
\notat{desired Regular Dextrous Workspace}{dRDW}{RDW$_d$}

\title{An efficient combined local and global search strategy for optimization of parallel kinematic mechanisms with joint limits and collision constraints}
\author[add1]{Durgesh Haribhau Salunkhe}
\author[add1,add2]{Guillaume Michel}
\author[add3]{Shivesh Kumar}
\author[add4]{Marcello Sanguineti}
\author[add1]{Damien Chablat}
\address[add1]{Laboratoire des Sciences du Numérique de Nantes (LS2N)}
\address[add2]{Centre Hospitalier Universitaire de Nantes}
\address[add3]{Deutsches Forschungszentrum f{\"u}r K{\"u}nstliche Intelligenz, Bremen}
\address[add4]{Universita degli studi di Genova}
\date{February 2022}

\begin{document}

\begin{abstract}
    The optimization of parallel kinematic manipulators (PKM) involve several constraints that are difficult to formalize, thus making optimal synthesis problem highly challenging. The presence of passive joint limits as well as the singularities and self-collisions lead to a complicated relation between the input and output parameters. In this article, a novel optimization methodology is proposed by combining a local search, Nelder-Mead algorithm, with global search methodologies such as low discrepancy distribution for faster and more efficient exploration of the optimization space. The effect of the dimension of the optimization problem and the different constraints are discussed to highlight the complexities of closed-loop kinematic chain optimization. The work also presents the approaches used to consider constraints for passive joint boundaries as well as singularities to avoid internal collisions in such mechanisms. The proposed algorithm can also optimize the length of the prismatic actuators and the constraints can be added in modular fashion, allowing to understand the impact of given criteria on the final result. The application of the presented approach is used to optimize two PKMs of different degrees of freedom.
\end{abstract}
\maketitle
 \section{Introduction}
\label{chapter:introduction}  
    \indent A \ac{PKM} is a closed-loop mechanism with multiple legs that are connected to the end-effector from the base. These mechanisms have seen recent rise in applications due to their high speed, high load and precision capacity in contrast to serial mechanisms \cite{Merlet_book}. Conversely, their design is more difficult because of singularities within their workspace and kinematic models that are difficult to calculate \cite{YDPatel, muller_singular_2019}. The first applications of these architectures were flight simulators with the Gough-Stewart platform \cite{uttm_gough} and pick-and-place robots with the Delta robot \cite{clavel_delta}.
    
    Due to their advantages, \ac{PKM} are used as sub mechanism modules in series-parallel hybrid robots in various fields such as humanoids, (THOR \cite{lee_design_2014}, LOLA \cite{lohmeier_modular_2006}, Charlie \cite{kuehn_system_2017}), exoskeletons \cite{kumar_design_2019, kumar2019modular}, haptic interface \cite{Delta_Haptic}, surgeries \cite{michel_new_2020}, and industrial applications \cite{dutta_sensorless_2019, udai_overlaid_2018}), see~\cite{KUMAR2020102367} for an extensive survey. \ac{PKM}s are also prominently employed in high speed industrial assembly lines, for example the DELTA + 1 DOF wrist robot \cite{brinker_comparative_2017}. Another important application of PKMs is the machining of parts, and they have been considered for milling operations as well as high speed machining tasks \cite{bernard_multiobjective_2011, ma_design_2018, wenger_orthoglide_2000}.\\
    Given the wide applications, the design of \ac{PKM} must meet user needs and process constraints. These needs may be related to the mobility of the robot, the size of its workspace, its movement accuracy, its dynamic performance, and its stiffness. Numerous performance indices have been defined to meet these requirements, which can be used in optimization problems \cite{gosselin_global_1991, chiu_kinematic_1988, vaf_chablat_orthoglide}. These include the conditioning of the Jacobian matrix, velocity amplification factors and regular workspace shapes as discussed in \cite{chablat_interval_2007}
    
    \begin{figure}
        \centering
        \includegraphics[width = 0.8\textwidth]{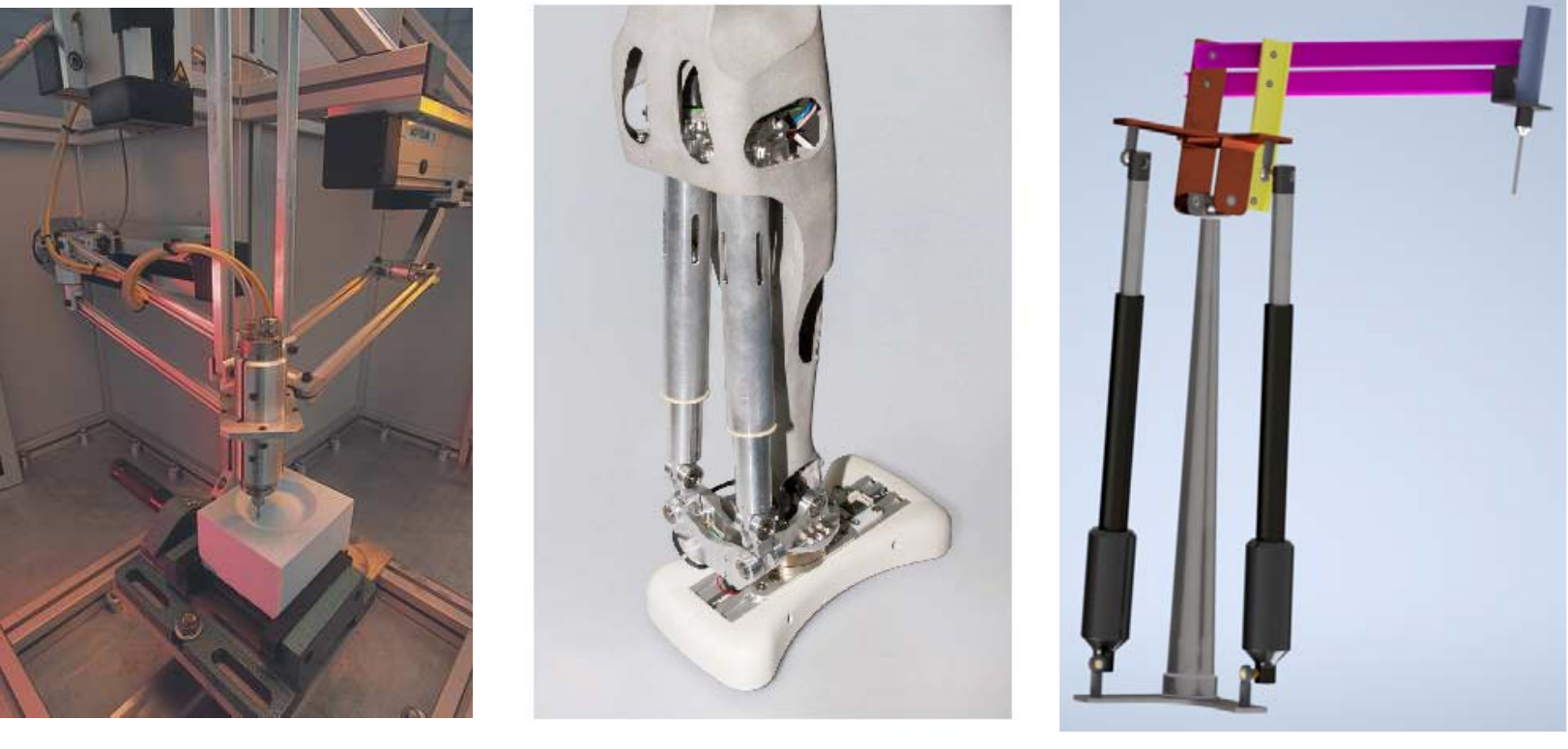}
        \caption{The parallel mechanisms used in different applications. From left: Orthoglide used in milling operations \cite{wenger_orthoglide_2000}, active ankle used in humanoids \cite{lenarcic_kinematic_2019}, RCM mechanism used in surgery \cite{michel_new_2020}}
        \label{fig:pkm_apps}
    \end{figure}
    
    In the past, several optimization methods were proposed for mechanism synthesis. Some of them utilize the mathematical formulation of the objective function in order to implement the gradient descent method \cite{germain_optimal_2013}. Where the objective function is not available in closed form and/or one cannot exploit gradient-based algorithms, numerical approaches and evolutionary algorithms were extensively implemented. Among them, we mention \ac{DE} \cite{saadatzi_multi-objective_2011} and Genetic Algorithms (GA) \cite{gallant_synthesis_2002} for single objective optimization and Branch and Prune \cite{caro_branch_2014}, Interval based analysis \cite{chablat_interval_2007} and \ac{NSGA-II} \cite{kucuk_dexterity_2009, ganesh_design_2020, hassan_modeling_2017, muralidharan_methods_2020} for \ac{MOO}, in which the theory of genetic evolution is implemented. Other evolutionary algorithms that have been implemented are \ac{PSO} \cite{zhang_design_2020} and \ac{MOEA/D}, which are claimed to be superior to \ac{NSGA-II} \cite{leal-naranjo_comparison_2019}. In general, the above-mentioned algorithms are computationally expensive, and their efficiency highly depends on the population size. Also, only a guess of the required set of initial population is available for convergence in the global search. The computational time increases considerably with increasing population size and thus limits the application of such methodologies in case of computationally expensive objective functions and also limits the number of constraints that can be implemented. \\ A recent work in mechanism design optimization is co-optimization with the motion trajectories \cite{ha_computational_2018}. In this approach, the design parameters and the motion equations are represented implicitly and efficient algorithms are used to explore the implicitly defined manifold. This type of methodology utilizes all the advantages of expressing the problem as an implicit function, but it is not always possible to do the same.\\
    
    To reduce the computational cost of optimizing a mechanism, a local search method can be implemented. To avoid the solution converging  to a local optimum, different methodologies are employed to combine local optimization methodologies with global searches \cite{durand_combined_1999, elleithy_globalization_2008, luersen_globalized_2004, niegodajew_power_2020}. Most of the literature presented above focuses primarily upon the problem formulation and use the existing methodology as an optimization tool. A deeper analysis into the implementation of the optimization algorithm for \ac{PKM} provides more flexibility and capability to handle different constraints efficiently. The geometrical method of proposing the next best solution in Nelder-Mead algorithm is best suited for mechanism optimization as the properties of the mechanisms are influenced by the lengths of the links and the exploration of design space in Nelder-Mead approach is very relevant to convergence to optimized parameters.\\
  
    In this work, we present a new design optimization methodology that can adapt to constraints involving internal collisions along with the physical joint limits and physical stroke of the actuator and classical criteria such as the condition number or the velocity amplification factor (VAF). We propose a fast local search algorithm, i.e. the Nelder-Mead algorithm, coupled with a global search procedure. A novel method is proposed to the local search by using different initialization in enabling one to compare results from a greater span in the optimization space. This method allows moving towards a global optimum faster, even for mechanisms that have computationally expensive objective functions. The overall output of the work is an accelerated  general algorithm for \ac{PKM} design optimization which is flexible with respect to the definition of the objective function as well as is modular and adaptive to any constraints. Two different \ac{PKM}s used in different applications are optimized using the proposed method to illustrate the advantage in terms of flexibility of the methodology.\\
    
    The paper is organized in the following way: Section \ref{chapter:optimization_elements} discusses the objective functions and constraints relevant to \ac{PKM} design. It highlights the importance of choosing proper constraints for mechanisms with prismatic joint. Section \ref{chapter:optimization_algorithm} details the optimization methodology that combines local and global searches and illustrates the novelty in accelerating the local search. Section \ref{chapter:implementation} provides the examples for the two \ac{PKM} design optimization with different objective functions and constraints and their corresponding optimized parameters. In Section \ref{chapter:conclusions}, some conclusions are presented, along with a few pointers to future work.

 \section{Design considerations in PKM optimization}
\label{chapter:optimization_elements}

\noindent In the parallel kinematic mechanism design the following choices have to be made: 
   \begin{enumerate}
       \item Architecture of the manipulator (e.g: 3R\underline{R}R(Revolute-Revolute(actuated)-Revolute), 3R\underline{P}R(Revolute-Prismatic(actuated)-Revolute) etc.)
       \item Type of joints: different combinations of joints to achieve the same \ac{dof} (e.g: U\underline{P}S(Universal-Prismatic(actuated)-Spherical), \underline{R}US, \underline{R}RPS)
       \item Pose of the joints: where to place and how to place a particular joint's frame? 
   \end{enumerate}
   Making a particular choice is non-trivial, especially because of its effect on the workspace, the kinematic solutions and the size of the mechanism. Another interesting challenge is that the same architecture can be used to perform different tasks with either kinematic or dynamic constraints, and thus have to be optimized accordingly. The following subsections elaborate on the common objective functions and constraints involved in mechanism optimization to motivate the choice of the algorithm.\\
   
    \subsection{Objective function}
    \noindent It is important to evaluate the quality of the motion performed while designing a manipulator with kinematic characteristics. The quality indices widely used in the past are the conditioning number \cite{gosselin_global_1991} and the manipulability ellipsoid \cite{chiu_kinematic_1988}. The feasible workspace and the global quality of the manipulator are directly related in the presented case, and thus can be implemented together with appropriate weights. 
    
     \subsubsection{Workspace of the manipulator}
     \noindent This work considers a \ac{RDW} without singularity which is an n-dimensional sphere in the n-dimensional output space and the center of the workspace is the home configuration (i.e, where all the actuator values are zero). To allow the implementation of the mechanism for multipurpose applications, the required workspace is not treated as a constraint. Instead, the algorithm tries to achieve maximum feasible workspace in the desired RDW ($RDW_d$) \cite{chablat_interval_2007}. \\
     
   In contemporary, the concept of {\em safe working zone} for parallel manipulators has been introduced in \cite{sfw_sandipan_2014} where a feasible workspace is free of singularities and internal link collisions and satisfies passive joint limits. This work considers only the collision of actuating prismatic joints, as the rest of the links can be redesigned to counter the resulting collision issues, if any. The context of \ac{FS} in this literature relates to the set of all points in the discretized \ac{op_space} such that:
   \begin{enumerate}
       \item They are non-singular configurations
       \item Respect passive joint limits
       \item For any postures, there is no internal collision between the actuators and the moving platform
   \end{enumerate}
   
   \subsubsection{Quality of the manipulator}
    The \ac{cnum} was introduced in \cite{gosselin_global_1991} to quantify the quality of motion. It is defined as the value of the asymptotic worst-case relative change in the output for a relative change in the input, and is used to measure how sensitive the output is to changes in the input. The geometrical interpretation of \ac{cnum} is the quantity proportional to the eccentricity of the ellipsoid, giving information about the ease of travel in a particular direction from a current end effector pose. When the \ac{cnum} is equal to 1, we have a sphere, and it corresponds to the {\em isotropic configuration}. The value of \ac{cnum} ranges from 1 to $\infty$ and so its inverse,  $\kappa^{-1}$, is used for bounded values and is given by (\ref{eq:conditioning_number}), where $\sigma$ are the singular values of the Jacobian matrix, $\mathbf{J}$.\\
    \begin{equation}
        \kappa^{-1} = \dfrac{\sigma_{min}}{\sigma_{max}}, \hspace{3mm} \kappa^{-1} \in [0, 1]
        \label{eq:conditioning_number}
    \end{equation}
    
    The conditioning number suffers from dimensional non-homogeneity of the Jacobian matrix and is not suitable for manipulators with both translational and rotational movements \cite{hunt_review_2003}. This is an important issue to consider while implementing the proposed optimization methodology for a general manipulator. The manipulators presented in Section \ref{chapter:implementation} have only rotational, \ac{dof} and so the inverse of the conditioning number is chosen as the quality index. A \changes{{\em\ac{gqi}} (GCI)}, the mean of summation of the values of \ac{qi} over the regular dextrous workspace, is defined as follows,
    \begin{equation}
        \kappa^{-1}_{g} := \dfrac{\sum\limits_{1}^{RDW_d}\kappa^{-1}}{RDW_d}
        \label{eq:gqi}
    \end{equation}
    
    \subsection{Constraints}
    \label{subsection:constraints}
   Parallel Kinematic Manipulators (PKMs) have three distinct features from a serial chain:
   \begin{enumerate}
       \item singularity inside the workspace where the control of the end-effector is lost;
       \item passive joints whose orientation can be calculated but not controlled explicitly;
       \item multiple legs - serial chains connecting the end-effector with the base.
   \end{enumerate}
   These \changes{three} features are of great importance as they affect the workspace of the manipulator as well as the nature of the motion. So, we take note that the passive joint limits and avoiding internal collisions among different legs of the \ac{PKM} are two important constraints to be implemented in our optimization problem. 
   
   \subsubsection{Non-singular constraint}
   \label{subsection:singular_constraint} 
\begin{figure}[ht]
    \centering
    		\centering
    		\includegraphics[width=4.5 in,height = 2 in]{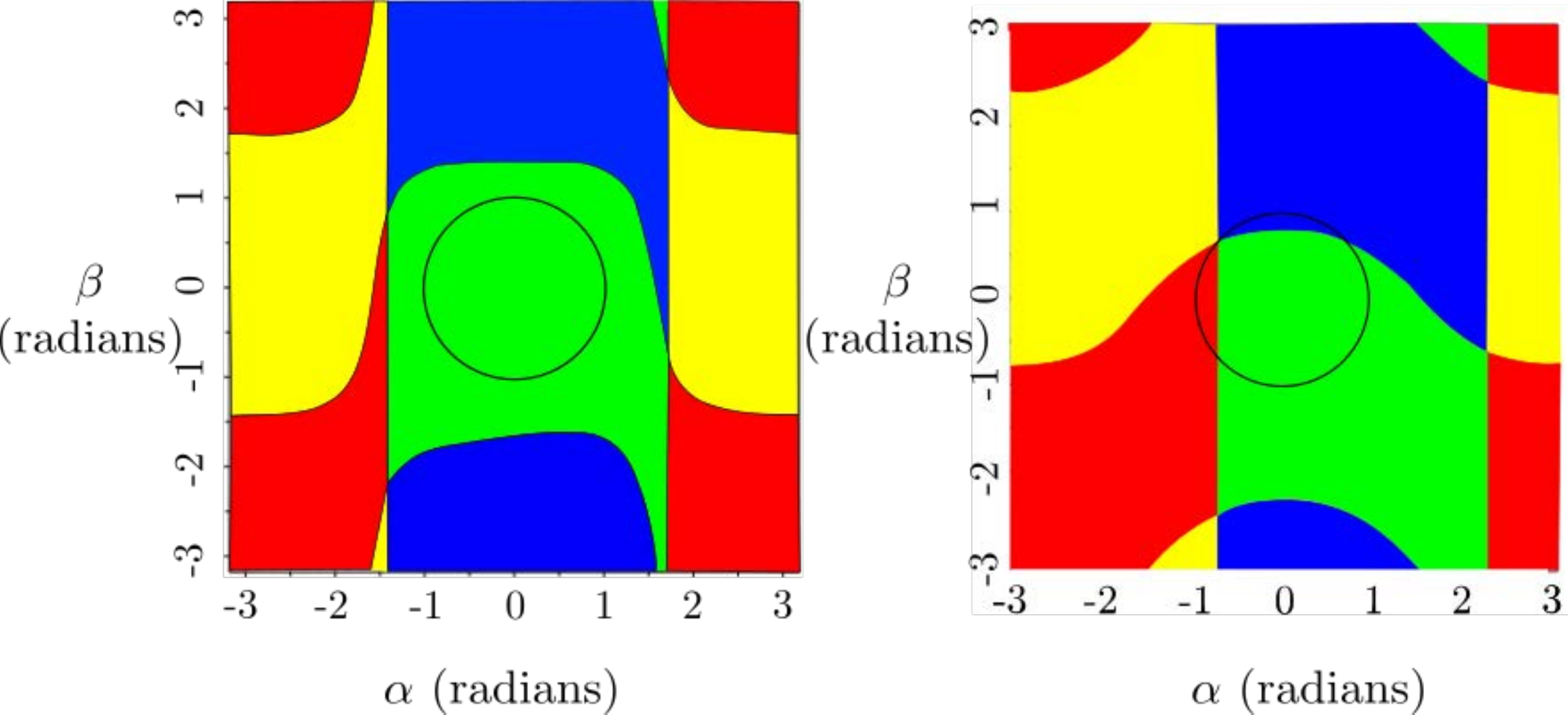}
    	\caption{The {\em aspects} in the workspace with different parameters for the 2UPS-1U mechanism from \cite{michel_new_2020, chablat_workspace_2021}}
    	\label{fig:RDW_aspects}
    \end{figure}

   The discretized output space (\ac{op_space}) of the parallel mechanism is separated by the singularity surfaces, thus resulting into several connected regions also called {\em aspects} \cite{chablat_working_1998}. As it is not possible to travel from one {\em aspect} to another, it is important that the desired RDW ($RDW_d$) lies in a single {\em aspect}. As an example, Figure \ref{fig:RDW_aspects} represents the output space for a mechanism with 2 orientation dofs and the black circle is the desired workspace. Figure \ref{fig:RDW_aspects}(left) illustrates a valid set of parameters whereas figure \ref{fig:RDW_aspects}(right) corresponds to a non-valid architecture of the mechanism because it consists of multiple connected regions, then there are multiple aspects, and thus we cannot travel to every configuration in the RDW.
    
    \subsubsection{Passive joints}
    \label{subsection:passive joints} 
    The architecture of \ac{PKM} is such that there are multiple passive joints in each leg. These passive joints are of prime importance in deciding the nature of the degree of freedom as well as the singular configuration. The joint positions as well as their limits are critical for the analysis. The passive joints can be either prismatic, revolute or a higher pair of joints. It is considerably easy to specify the limits in case of prismatic or revolute joint, but it may not be easy when we use a universal joint or a spherical joint. Implementing passive joint limits in the optimization process allows achieving a practical result for the design and gives more clarity about the feasible workspace.
    
    \subsubsection{Link collisions}
    \label{subsection:link_collision} 
    As \ac{PKM} consists of multiple serial chains attached to the end effector from the base, internal collisions are an important aspect of workspace analysis and synthesis as they strongly affect the workspace and other kinematic properties. The theoretical workspace is generally different from the practical realization, because of the mechanical joint limits and link collisions in the mechanism \cite{chablat_moveability_1998}. Analysis of such self collisions is critical in parallel manipulators. The implementation of link collisions as a constraint is even more complicated since it depends not only on the architecture of the serial legs but also on the design of the links as well as their assembly. Different approaches have been used to calculate the collision between the links. The work reported in \cite{Merlet_selfadd1, Merlet_selfadd2} uses a common normal to determine the distances between two links, while a more modern approach is to use Computer Aided Design (CAD) \cite{DANAEI2017230, fcl_cite}. The presented work only considers the collision between the actuators, and thus the method suggested in \cite{Merlet_selfadd1, Merlet_selfadd2} suffices.
    
    \subsubsection{Feasible actuator range: constraint for mechanisms with prismatic actuators}
    \label{subsection:feasible_range} 
\begin{figure}[ht]
    \centering    		\includegraphics[width=0.5\textwidth]{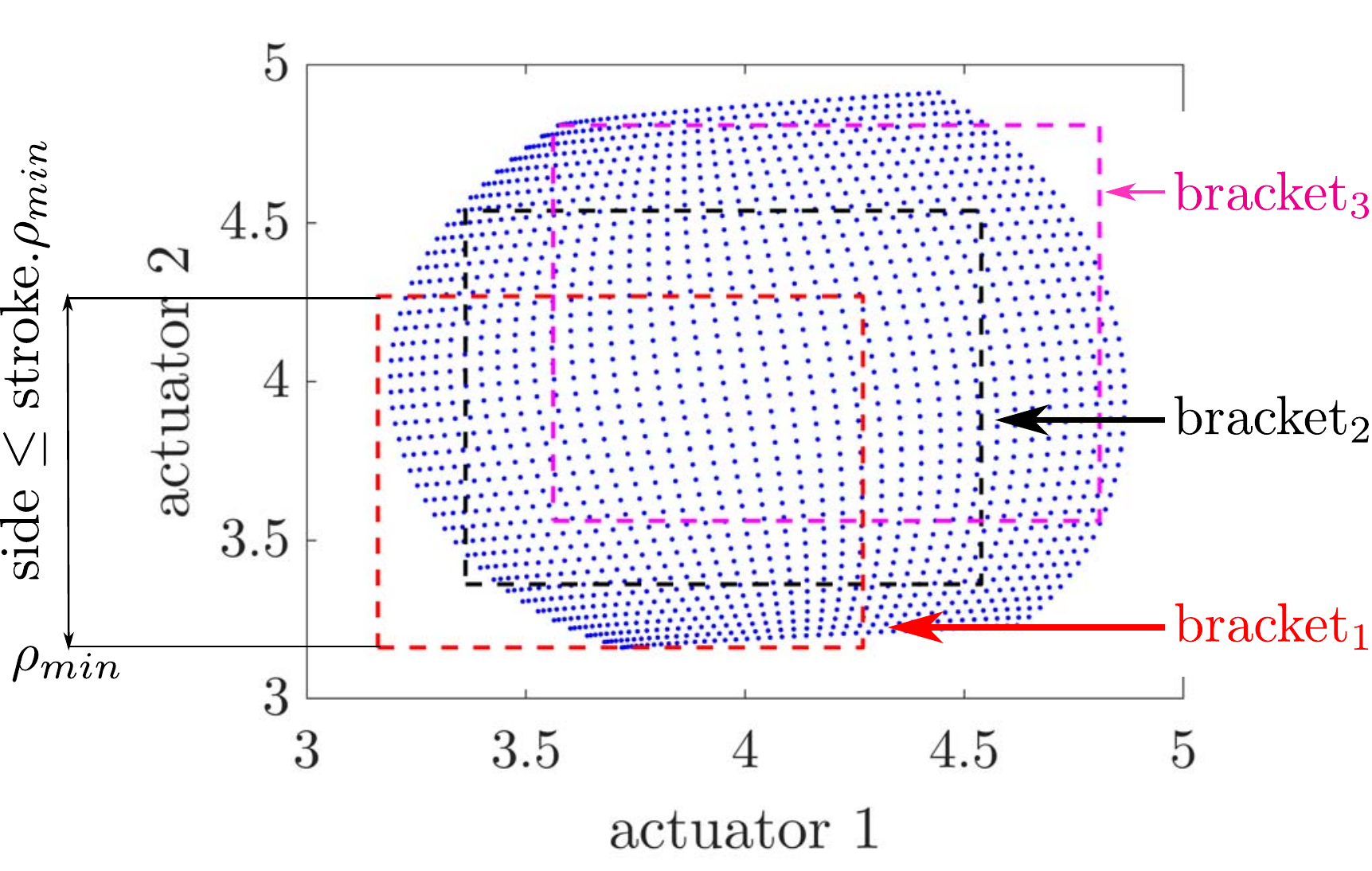}
    \captioncomment{Different search brackets within the actuator space (\ac{ip_space})}{The dots correspond to the pair of lengths of actuators for a configuration in RDW.}    	
    \label{fig:actuator_search}
    \end{figure}

\begin{figure}[ht]
    \centering
    	\begin{subfigure}{0.3\textwidth}
    		\centering
    		\includegraphics[width=1.8in,height = 1.5 in]{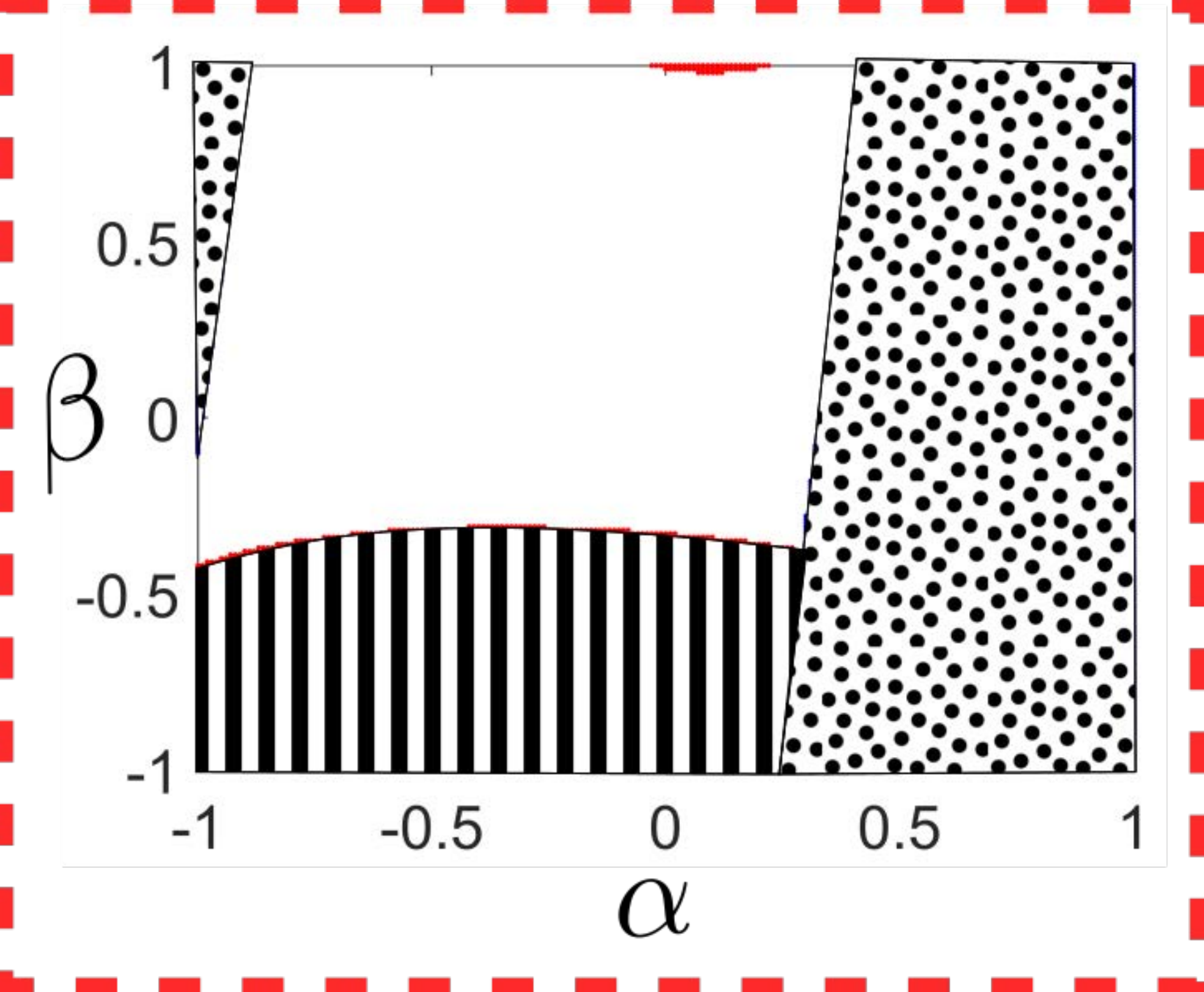}
    		\caption{Feasible workspace (white) when bracket 1 in figure \ref{fig:actuator_search} is implemented}
    		\label{fig:valid_plot_red}
    	\end{subfigure}
    	~
    	\begin{subfigure}{0.3\textwidth}
    		\centering
    		\includegraphics[width=1.8in,height = 1.5 in]{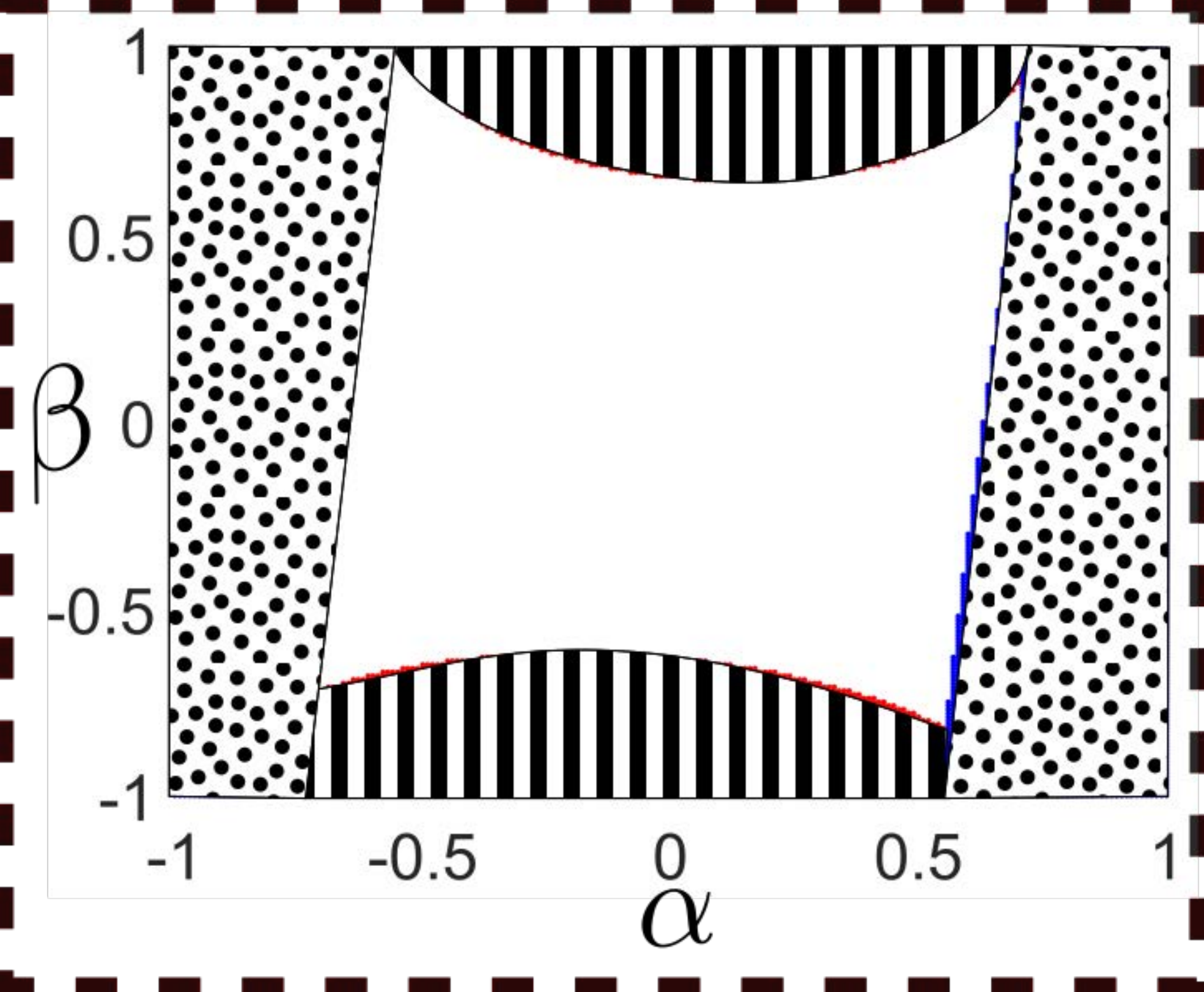}
    		\caption{Feasible workspace (white) when bracket 2 in figure \ref{fig:actuator_search} is implemented}
    		\label{fig:valid_plot_black}
    	\end{subfigure}
    	~
    	\begin{subfigure}{0.3\textwidth}
    		\centering
    		\includegraphics[width=1.8in,height = 1.5 in]{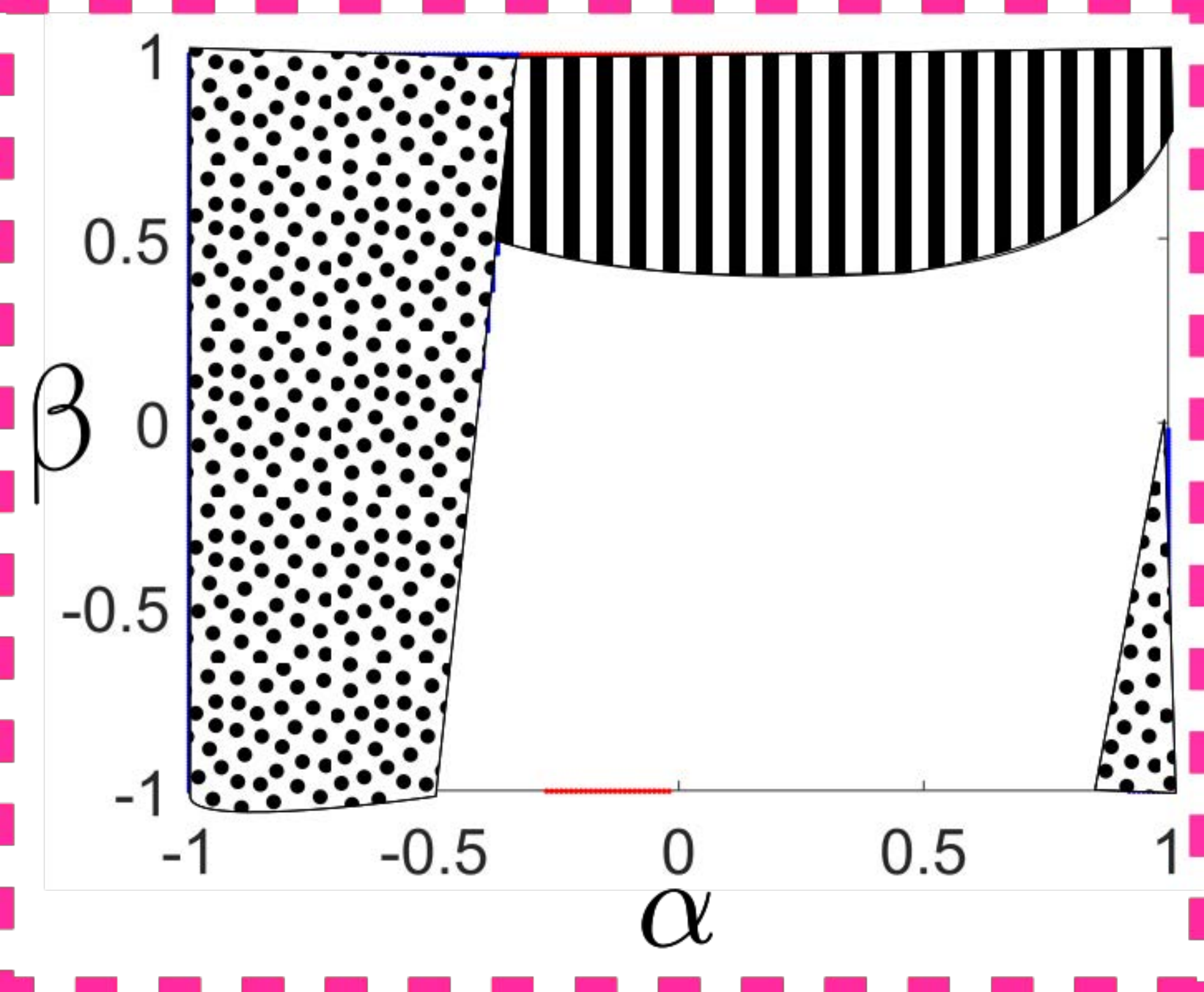}
    		\caption{Feasible workspace (white), bracket 3 in figure\ref{fig:actuator_search} is implemented}
    		\label{fig:valid_plot_magenta}
    	\end{subfigure}
    	\captioncomment{Comparison of feasible workspace (white space) within the $RDW_d$ for different search brackets and a specific mechanism (2UPS-1U)}{The striped and dotted part represent the violation due to actuator lengths of first and second leg, respectively.}
    	\label{fig:plot_valid_compare}
    \end{figure}

    Another important constraint while designing a \ac{PKM} is the active joint ranges. This constraint is specifically relevant to the mechanisms with prismatic joints as actuators. The aim is to implement a constraint on the actuators to be chosen in order to maximize the points in $\mathcal{F} \bigcap RDW_d$. Generally, a prismatic joint is expressed as a constraint with a certain minimum and maximum range and with a constraint on the ratio between the length in completely actuated state and its default length:
    \begin{equation}
        \rho_{min} \leq \rho \leq \rho_{max}
        \label{eqn:static_rho}
    \end{equation} 
    \begin{equation}
        \rho_{max} \leq \text{stroke}\cdot \rho_{min}, \hspace{3mm} stroke \in [1, 2]
        \label{eqn:stroke_rho}
    \end{equation}
    \begin{figure}[H]
        \centering
        \includegraphics[width=0.7\textwidth]{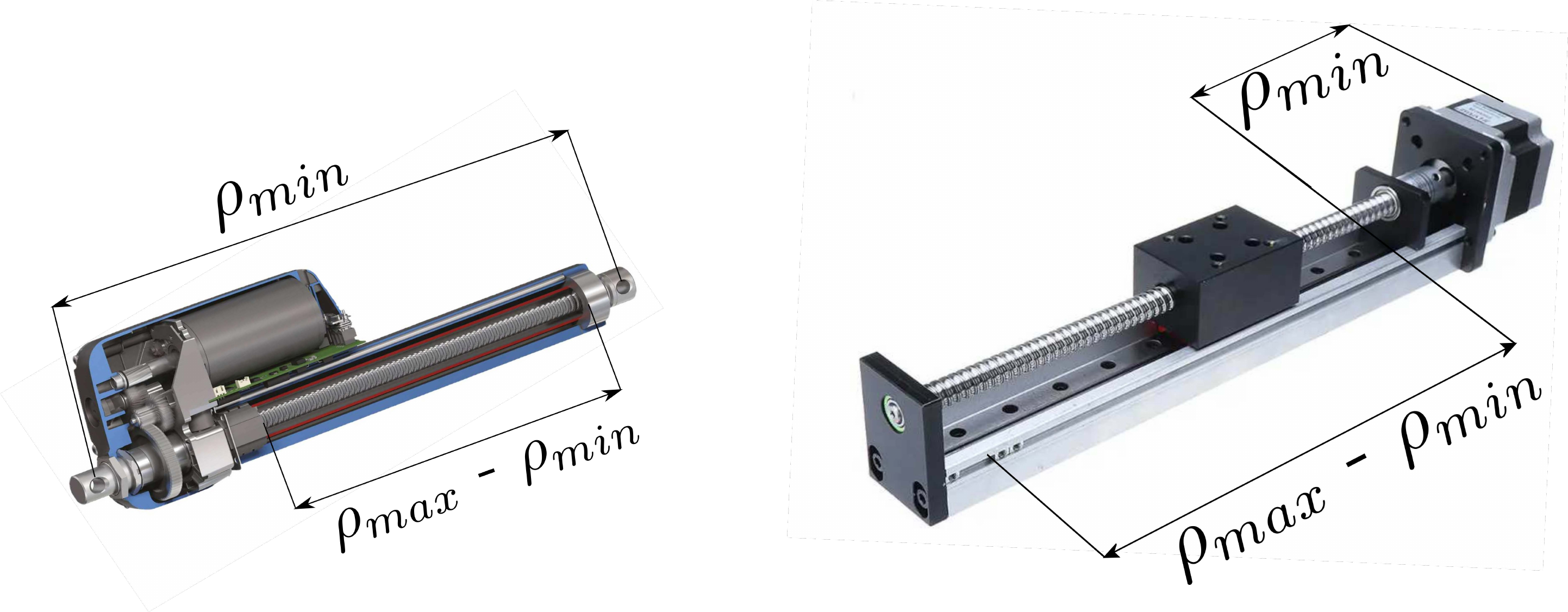}
        \caption{\changes{The industrial prismatic joints and the relation between $\rho_{min}$ and $\rho_{max}$. Source: Hanpose linear actuator HPV5 SFU1204,  \url{www.pngegg.com/en/png-mckkp}.}}
        \label{fig:linear_actuators}
    \end{figure}
    Equation \ref{eqn:stroke_rho} comes from the physical structure of general prismatic joints. If the unextended length of the actuator is $\rho_{min}$, then it is not practical for common prismatic joints to extend beyond their original length ($\rho_{max} < 2\cdot\rho_{min}$) as explained in figure \ref{fig:linear_actuators}. The novelty in expression of the actuator range in the present work is that we do not have a static value as a limit as mentioned in \changes{Equation} \ref{eqn:static_rho}, i.e, we express the constraint only in terms of the stroke ratio defined in \changes{Equation} \ref{eqn:stroke_rho}. This allows us to choose the best actuator ranges to maximize the feasible workspace without putting any constraint on the minimum or maximum size of the prismatic joint. This is illustrated in figures \ref{fig:actuator_search} and \ref{fig:plot_valid_compare} which introduce an example for a 2 dof 2UPS-1U orientation mechanism from \cite{michel_new_2020}. The points in the dotted space in figure \ref{fig:actuator_search} are pairs of values corresponding to the actuator lengths in a feasible configuration. The aim is to search for an optimized bracket, [$\rho_{min}, \rho_{max}$], i.e, a bracket that includes as many blue points as possible with the constraint that the side of the square does not exceed a given proportion with respect to its minimum value. \\
    
    The algorithm \ref{algo:best_rho} explains the method used to get the optimized bracket for the actuators. After discretizing the $RDW_d$, we get the set of all valid points belonging to $\mathcal{F}$. Upon calculating the values for actuator length at each point, the minimum $\rho_{min}$ and maximum $\rho_{max}$ value for the actuator is obtained. The input of the algorithm is a \textit{n} x 3 matrix for the \textit{n} valid points, with columns corresponding to the actuator lengths and the evaluation at that point. If the ratio of maximum value to minimum value of the actuator length respects the stroke ratio, then the algorithm returns the actuator range without alteration. Otherwise, a bracket of [$\rho_{min}$, stroke.$\rho_{min}$] is generated and the values of the actuator lengths for each point in the set of valid points is checked against the bracket and the number of points satisfying the bracket is stored. This process is repeated by incrementing the $\rho_{min}$ till the value of stroke.$\rho_{min}$ is lower than $\rho_{max}$. The algorithm returns the optimized actuator lengths along with the corresponding evaluation of the objective function for given parameters.
  
    \begin{algorithm}[H]
    	\SetAlgoLined
    	\KwResult{evaluation, \textit{e}, and corresponding range of actuators, $\mathbf{\rho_{range}}$}
    	$\mathbf{input}$ $\rightarrow$ $\mathbf{valid\_points}$ that satisfy all the constraints\;
    	$\mathbf{\rho_{range}}$ = [min($\mathbf{\rho_{vec}}$), max($\mathbf{\rho_{vec}}$)] = [$\rho_{min}$, $\rho_{max}$]\;
    	stroke : The maximum allowable stroke ratio, see Eq. (\ref{eqn:stroke_rho})\;
    	steps : number of brackets used (see figure \ref{fig:actuator_search})\;
    	\texttt{\\}
    	Checking for feasible set of the actuators\;
    	\eIf{$\rho_{max} \geq stroke\cdot\rho_{min}$}{
    		\For{$\rho_{lower}$ \textbf{from} $\rho_{min}$ \textbf{to} $\dfrac{\rho_{max}}{stroke}$ \textbf{by} steps}{
    			e = 0\;
    			\For{n \textbf{from} $1$ \textbf{to} length($\mathbf{valid\_points}$)}
    			{
    				\If{$\rho_1, \rho_2 \geq \rho_{lower}$ \textbf{and} $\rho_1, \rho_2 \leq$ stroke$\cdot\rho_{lower}$}{
    					j = j+1 \Comment{Incrementing the number of feasible points}\;
    					e = e + $\mathbf{valid\_points}$[n, 3]\;
    				}
    			}
    			$\mathbf{eval\_vector}$[k]= [e, $\rho_{min}$, j]\;
    			k = k + 1\;
    		}
    		[$e_1$, $\rho_{min}$, $i_1$] = max($\mathbf{eval\_vector}$, 1)\;
    		e = $\mathbf{eval\_vector}$[$i_1$][1]\;
    		$\rho_{min}$ = $\mathbf{eval\_vector}$[$i_1$][2]\;
    		$\mathbf{\rho_{range}}$ = [$\rho_{min}$, stroke.$\rho_{min}$]\;
    		\texttt{\\}
    	}
    	{e = max($\mathbf{valid\_points}$[3])\;
    		$\mathbf{\rho_{range}}$ = [min($\mathbf{\rho_{vec}}$), max($\mathbf{\rho_{vec}}$)]}
    	\Return $e$,  $\mathbf{\rho_{range}}$
    	\caption{Implementation for choosing the best actuator range}
    	\label{algo:best_rho}
    \end{algorithm}
    
    \subsubsection{Implementation of constraints and evaluation function}
    \label{subsubsection:mech_implement}
    Algorithm \ref{algo:evaluation} illustrates the methodology used to evaluate a given set of parameters. The optimization space is discretized, and each point is evaluated for the constraints. Some constraints are implemented strictly, in the sense that if even one point in the $RDW_d$ violates the constraint, then we discard the given set of parameters as an invalid solution. The singularity constraint is a strict constraint in the present algorithm, in the sense that if the singularity curve intersects with even one point of $RDW_d$, the evaluation for given parameters is negative. In  cases where the $RDW_d$ is singularity free, if all other constraints (e.g: passive joint limits, collision constraints) are satisfied at a particular point in the $RDW_d$ then it is rewarded by the corresponding $\kappa^{-1}$ value or if there is a violation of these constraints (except singularity) the point in $RDW_d$ is given 0 value. As we examine each point in the discretized workspace, the final evaluation is the cumulative value of $\kappa^{-1}$ over the workspace where all the constraints are satisfied. The rewarding strategy can be changed as per the designer's needs, and desirable weightage can be assigned to the constraints to achieve an optimized design for a specific requirement.\\
    The modularity of the algorithm with the constraints can also be observed in Algorithm \ref{algo:evaluation}. It can be seen that the constraints are completely independent of each other, allowing to activate, deactivate any constraint or add other constraints without requiring any change to the algorithm. This is especially useful for mechanism design, as it provides flexibility to experiment the effect of different constraints on the final feasible workspace. As each constraint can be designed individually to reward or penalize a particular set of parameters, the designer can have a blend of strict and non-strict constraints in the optimization. The designer can also identify which constraint stops the optimization and needs to be modified. 
    \begin{algorithm}[H]
    	\SetAlgoLined
    	\KwResult{evaluation at a given point in optimization space and the corresponding actuator lengths}
    	$\mathbf{input}$ $\rightarrow$ $\mathbf{v}$ \Comment{It is a n-dimension point in given n-dimension optimization space}\;
    	$x_i, i \in {1, .., n}$, \Comment{$i^{th}$ variable of the n-dimension optimization space}\;
    	$\rho_1$ and $\rho_2$ \Comment{actuator lengths at a given configuration}\;
    	e = 0 \Comment{Initialising the evaluation}\;
    	\For{$x_1$ \textbf{from} $x_{1min}$ \textbf{to} $x_{1max}$ \textbf{by} $interval_i$}
    	{  ... \Comment{Add loops as a function of the dimension of the space}\\
    		\For{$x_n$ \textbf{from} $x_{nmin}$ \textbf{to} $x_{nmax}$ \textbf{by} $interval_n$}
    		{
    			f($\mathbf{v}$) \Comment{function that solves IGS, collision distance and $\kappa^{-1}$}\;
    			[det($\mathbf{J}$),\,$\mathbf{q}_p$,\, $\rho_1$,\, $\rho_2$,\, $\kappa^{-1}$, $d_c$] = f($\mathbf{v}$)\;
    			f($\mathbf{v}$) returns the value of the determinant of Jacobian, the passive joint angle vector, $\mathbf{q}_p$, actuator lengths, [$\mathbf{\rho}_1, \mathbf{\rho}_2$], the inverse of the conditioning number, $\kappa^{-1}$ and the collision distance, $d_c$, between the actuators\;
    			
    			\Comment{\underline{1. Checking for singularity constraints}}\;
    			\eIf{det$(\mathbf{J})$ \textbf{is} 0}{e = -$\infty$\; break\;}{$reward$ = $\kappa^{-1}$}
    			
    			\Comment{\underline{2. Checking the passive joint limits}}\;
    			\For{i \textbf{from} 1 \textbf{to} length of $\mathbf{q}_p$}{\eIf{$q_{pi} \geq q_{pmax}$ \textbf{or} $q_{pi} \leq q_{pmin}$ }{$reward = 0$}{$reward$ = $\kappa^{-1}$}}
    			
    			\Comment{\underline{3. Checking for collision constraints}}\;
    			\If{$d_c \geq$ threshold}{$reward$ = 0}
    			e = e + reward\; 
    			$\mathbf{valid\_points}$[i] = [$\rho_{1}$, $\rho_{2}$, reward]\;
    		}
    		...\\
    	}
    	Implement the algorithm \ref{algo:best_rho}\;
    	\Return  $\mathbf{valid\_points}, e, \mathbf{\rho}_{1}, \mathbf{\rho}_{2}$
    	\caption{Method to calculate the evaluation and $\mathbf{\rho}_{range}$ for a set of parameters}
    	\label{algo:evaluation}
    \end{algorithm}  
 
 \section{Proposed Algorithm for Mechanism Optimization}
\label{chapter:optimization_algorithm}
In this section, the complete optimization methodology is illustrated. Recalling from the previous sections, the aim is to implement an algorithm that is capable of handling the non-smooth objective functions as well as the constraints related to PKM design. This section is divided in three subsections detailing the local search, the global search and the strategy used to couple them both for faster and more efficient solutions, respectively.
\subsection{Local search algorithm: The Nelder-Mead (NM) algorithm}
\label{section:local_search}
    The Nelder-Mead algorithm is a derivative-free optimization algorithm proposed by John Nelder and Roger Mead \cite{nelder_simplex_1965}. It is also called the {\em downhill-simplex algorithm,} as it uses {\em simplexes} to search the space locally. In this section, we present the algorithm for a {\em single start}, which searches for the optimum solution in the local vicinity of the initial simplex. Later, we discuss the implementation of the algorithm in mechanism optimization and detail the method for extracting the best actuator ranges from the solution. The section is concluded with a summary of the algorithm and its implementation, highlighting few strengths and weaknesses of the same.\\
    
\begin{figure}[htbp]
    \centering
        \begin{subfigure}{0.5\textwidth}
        	\centering
            \includegraphics[width = \textwidth]{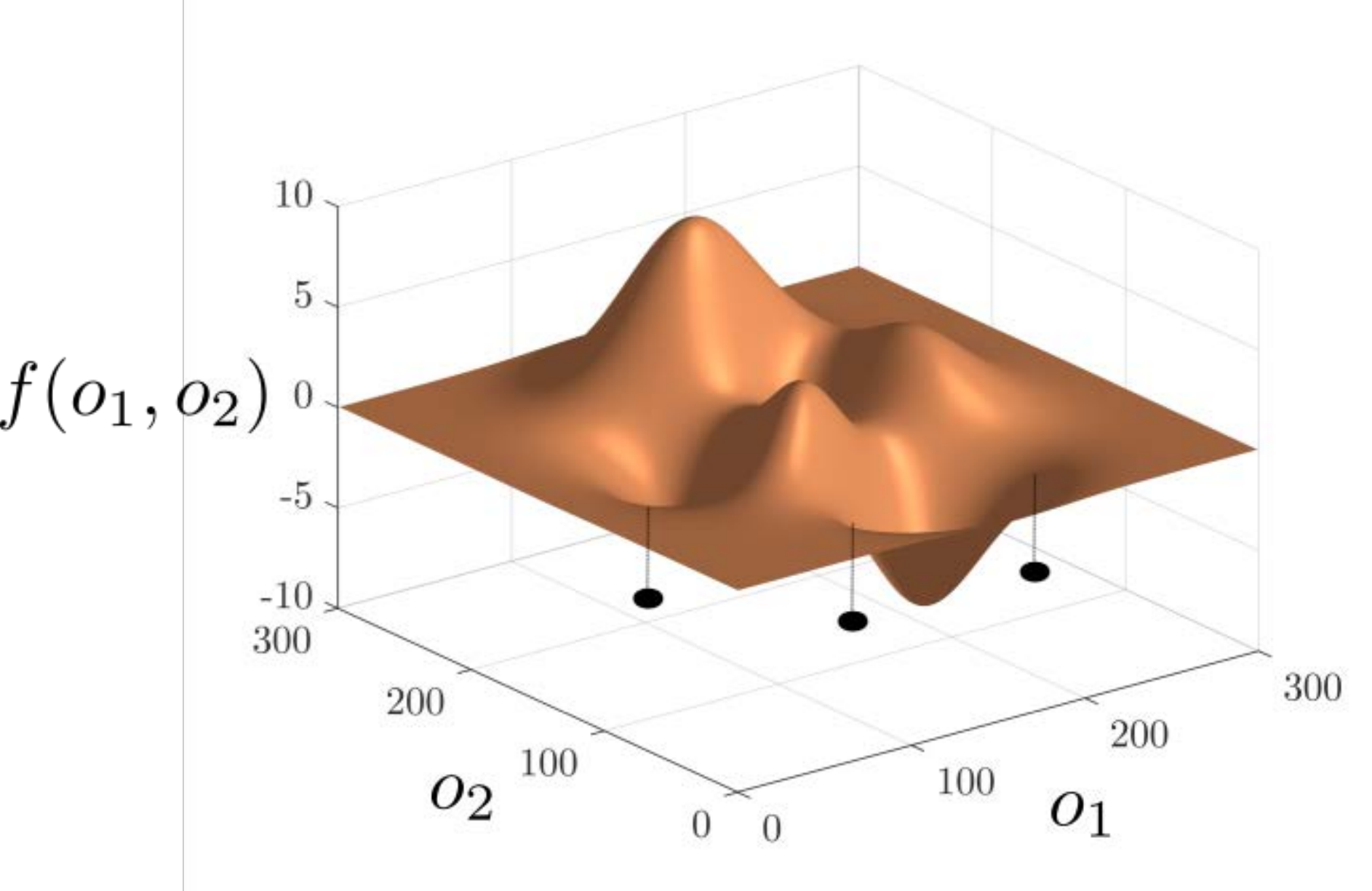}
            \caption{Example mapping with 2 optimization variables.}
            \label{fig:3d_peaks}
    	\end{subfigure}
    	\\
    	\begin{subfigure}{0.45\textwidth}
    		\centering
    		\includegraphics[width=\textwidth]{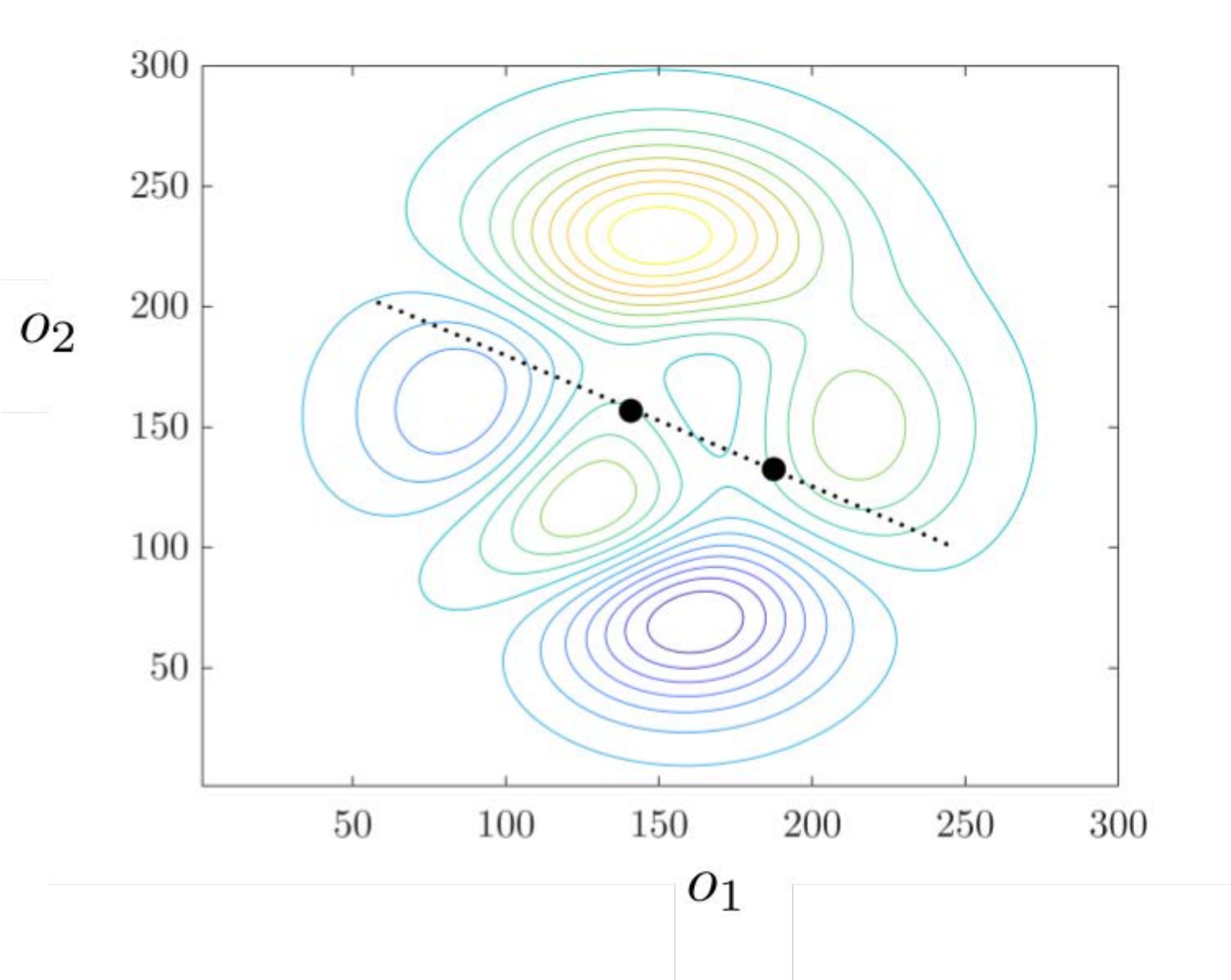}
    		\caption{The traversing in $\mathcal{O}$-space with 2 point simplex, we can explore the points on the line only}
    		\label{fig:contour_premat}
    	\end{subfigure}
    	~
    	\begin{subfigure}{0.45\textwidth}
    		\centering
    		\includegraphics[width=\textwidth]{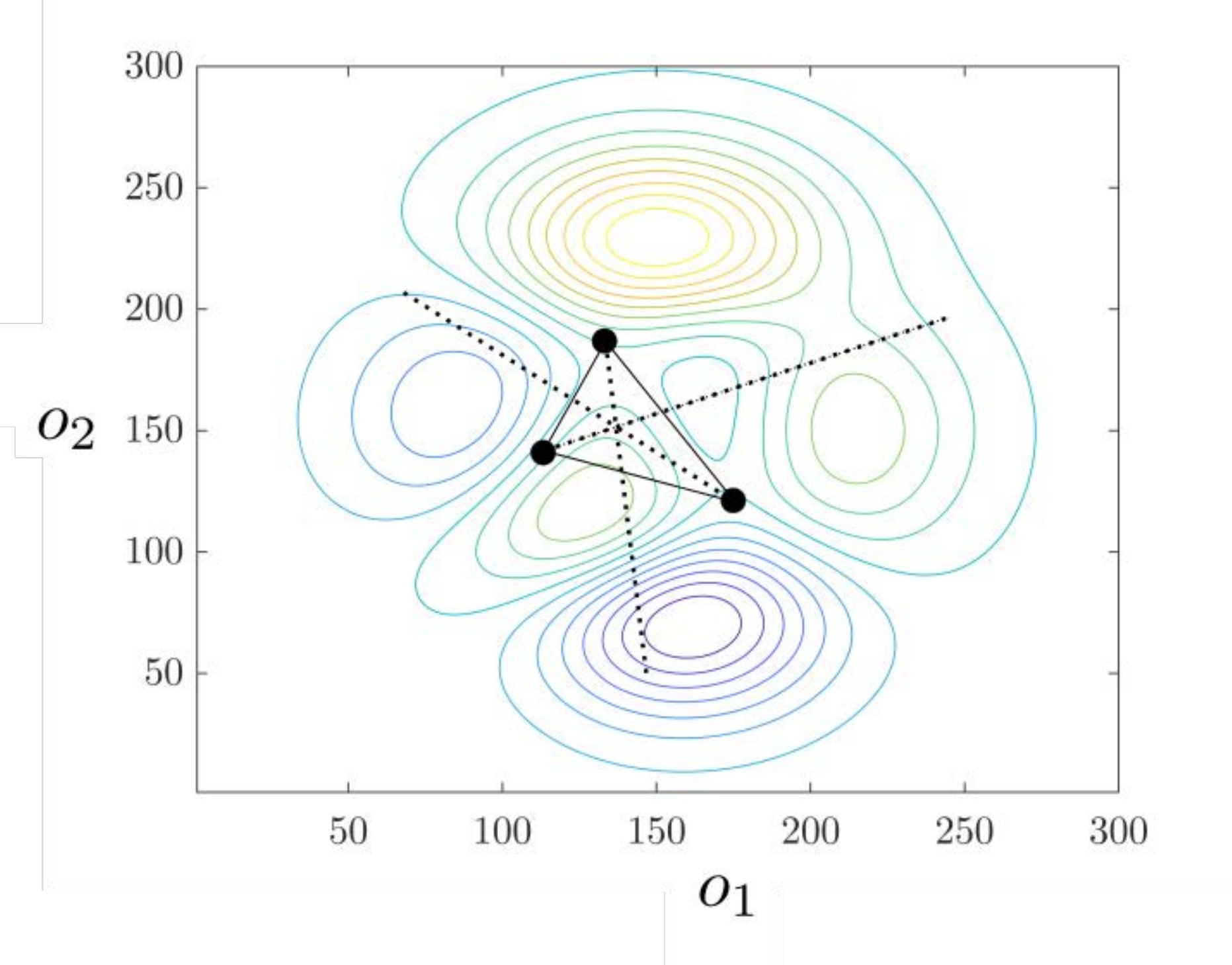}
    		\caption{The traversing in $\mathcal{O}$-space with 3 point simplex, allowing to explore the complete space}
    		\label{fig:contour_mat}
    	\end{subfigure}
    	\caption{Premature convergence when using a simplex of less than (n+1) points in n-dimensional $\mathcal{O}$-space. The example is taken from Matlab function [X, Y, Z] = peaks(25), MATLAB 2021}
    	\label{fig:contour}
    \end{figure}

    For a n-dimensional \ac{o_space}, we require a simplex of at least n+1 points in \ac{o_space} to avoid {\em premature convergence}. This can be explained with a simple graphics for 2-dimensional, \ac{o_space} as show in figure \ref{fig:contour}.\\

\begin{figure}[htbp]
        \centering
        \begin{subfigure}[t]{0.45\textwidth}
            \centering
            \includegraphics[width = \textwidth]{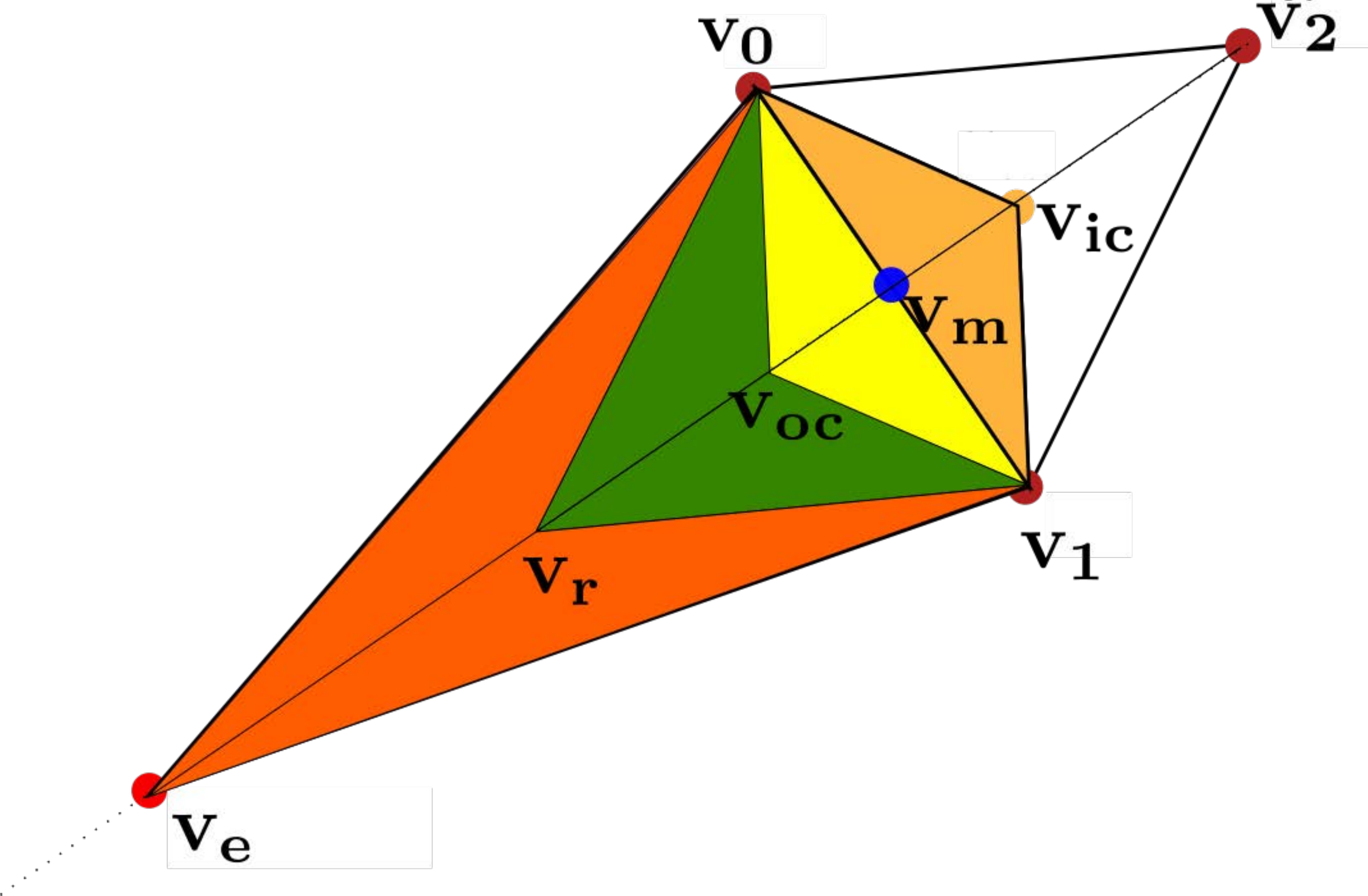}
        \caption{An example of an operation on a simplex (defined by $v_0, v_1, v_2$) in 2-dimensional $\mathcal{O}$}
        \label{fig:NM_example}
        \end{subfigure}
        ~
        \begin{subfigure}[t]{0.5\textwidth}
            \centering
            \includegraphics[width = \textwidth]{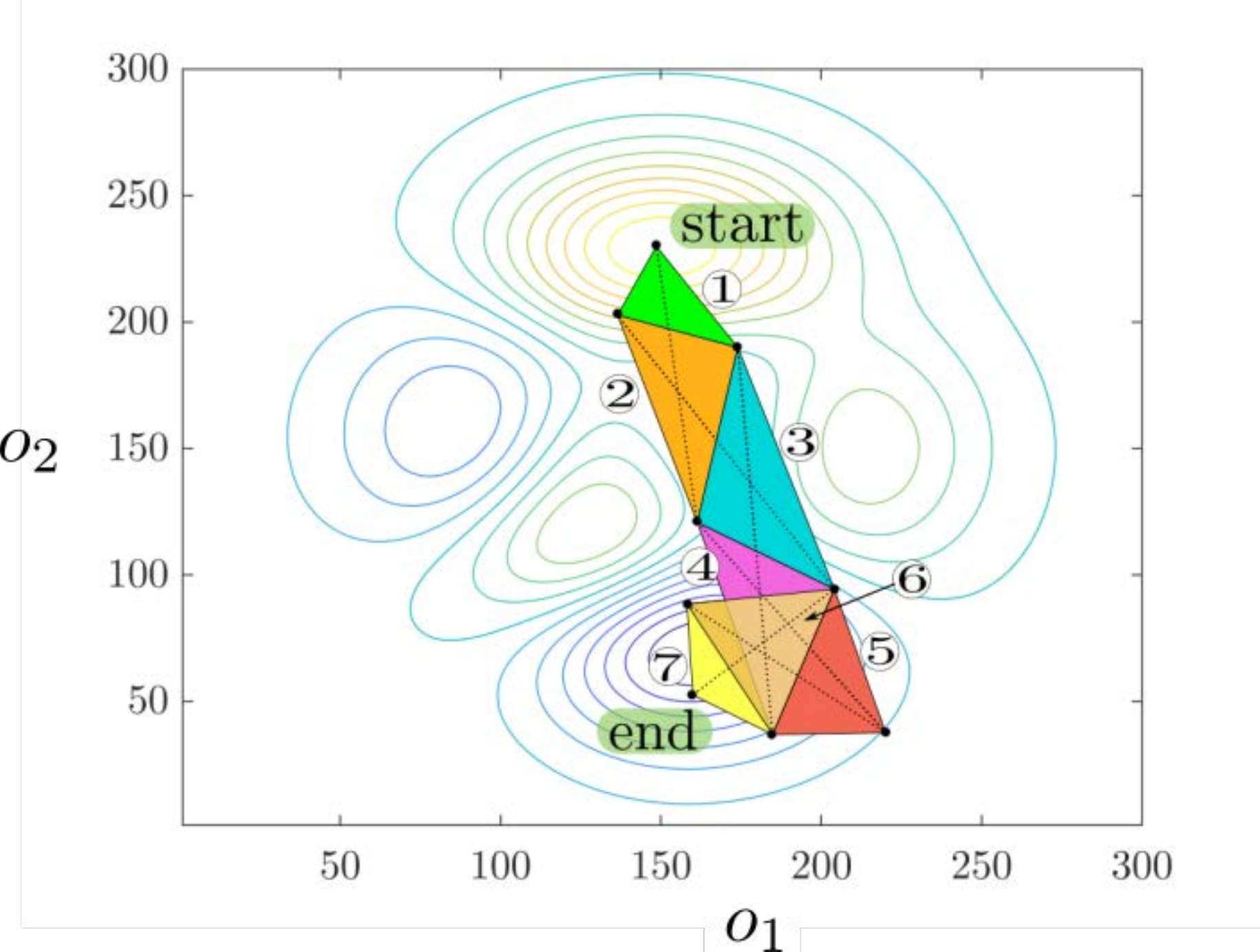}
        \caption{An example of the travel path of optimization in Nelder-Mead algorithm on the contour plot of peaks(25) function}
        \label{fig:NM_travel}
        \end{subfigure}
        \caption{The single start of the Nelder-Mead local search}
    \end{figure}

    The algorithm is initiated with a sorted simplex of n+1 points ($\mathbf{v_0}, \mathbf{v_1}, ... \mathbf{v_n}$) such that the objective function evaluated of the $i^{th}$ vertex has a value better than or equal to that of \changes{the} $(i+1)^{th}$ vertex. A mean point ($\mathbf{v_m}$) is calculated by excluding the worst point ($\mathbf{v_n}$):
    \begin{equation}
        \mathbf{v_m} := \dfrac{\sum\limits_{i = 0}^{n-1} \mathbf{v_i}}{n}
    \end{equation}
    The optimization algorithm then compares the mean point and searches for better points by geometrical operations termed as \point{i} reflection, \point{ii} expansion, \point{iii} inside contraction, \point{iv} outside contraction and \point{v} shrinkage. These operations are defined as follows:
    \begin{enumerate}
        \item Reflection ($\mathbf{v_r}$) :  \begin{equation}
        \mathbf{v_r} = \mathbf{v_m} + r\,(\mathbf{v_m} - \mathbf{v_n}), \hspace{3mm} r = \text{reflection coefficient}\, (r > 0) 
    \end{equation}
    \item Expansion ($\mathbf{v_e}$) : \begin{equation}
        \mathbf{v_e} = \mathbf{v_m} + e\,\mathbf{(v_r - v_m)}, \hspace{3mm} e = \text{expansion coefficient}\, (e > 1)
    \end{equation}
    \item Outside contraction ($\mathbf{v_{oc}}$) : \begin{equation}
        \mathbf{v_{oc}} = \mathbf{v_m} + k\,(\mathbf{v_m} - \mathbf{v_n}), \hspace{3mm} k = \text{contraction coefficient}\, (0< k < r)
    \end{equation}
    \item Inside contraction ($\mathbf{v_{ic}}$): \begin{equation}
        \mathbf{v_{ic}} = \mathbf{v_m} - k\,(\mathbf{v_m} - \mathbf{v_n}), \hspace{3mm} k = \text{contraction coefficient}
    \end{equation}
    \item Shrinkage: \begin{equation}
       \forall \, i \in [1, n] \hspace{3mm} \mathbf{v_{i}} = s\,.\mathbf{v_i}, \hspace{3mm} s := \text{shrinkage factor}\, (0< s < 1) 
    \end{equation}
    \end{enumerate}
    The new point ($\mathbf{v_{n}}$) introduced in the simplex depends on the evaluation of the $\mathbf{v_r}$, $\mathbf{v_e}$, $\mathbf{v_{oc}}$ and $\mathbf{v_{ic}}$ (see Algorithm \ref{algo:NM_single_op}). The operation is continued \changes{until} the stopping criteria \changes{are} reached. The simplex stops if it shrinks below a certain value, $\epsilon_1$ and the evaluations of every vertex of the shrunk simplex vary by a maximum threshold $\epsilon_2$. The algorithm can be stopped by limiting the number of iterations, too. The stopping criteria was presented in Algorithm \ref{algo:stop_criteria} and the complete procedure for one start of the \ac{NM}-algorithm is given in Algorithm \ref{algo:NM_single_op}. An example of the operations in a 2-dimension optimization space, \ac{o_space}, is illustrated in figure \ref{fig:NM_example} to present the geometric nature of search of the \ac{o_space} in \ac{NM}-algorithm. In figure \ref{fig:NM_travel}, an example of the points explored during an optimization process is graphically represented.
    
    \begin{algorithm}[H]
        \SetAlgoLined
        \KwResult{Boolean for stopping condition}
         sorted simplex $\{\mathbf{v_0}, \mathbf{v_1}, \mathbf{v_2}, ..., \mathbf{v_{n-1}}, \mathbf{v_n}\}$\;
         evaluations \{{\em $e_0, e_1, e_2, ..., e_{n-1}, e_n$}\}\;
         maximum iteration = {\em max\_iter}... from algorithm \ref{algo:NM_single_op}\;
         iteration count = iter\;
         {\em $l_{ij}$} = $\norm{\mathbf{v_j} - \mathbf{v_i}}$... length of the side of simplex\;
         {\em $e_{ij}$} = $\vert${\em $e_i$} - {\em $e_j$}$\vert$\;
         \If{max({\em $l_{ij}$}) $\leq \epsilon_1$ $\&\& $ max({\em $e_{ij}$}) $\leq \epsilon_2$}{stop = 1\\break}
         \eIf{iter $\geq$ max\_iter}{stop = 1}{stop = 0}
         \caption{Stopping criteria for the NM algorithm}
        \label{algo:stop_criteria}
    \end{algorithm}
    
    \begin{algorithm}
        \SetAlgoLined
        \KwResult{Local minimum evaluation and the optimized parameters}
         initial sorted simplex $\{\mathbf{v_0}, \mathbf{v_1}, \mathbf{v_2}, ..., \mathbf{v_{n-1}}, \mathbf{v_n}\}$\;
         evaluations $\{e_0, e_1, e_2, ..., e_{n-1}, e_n\}$\;
         
         \While{stop = 0 }{
         calculate $\mathbf{v_m, v_r}$ and $e_r$\;
          \uIf{($e_n < e_r < e_0$)}{
           $\mathbf{v_{n}}$ = $\mathbf{v_r}$\;
           }
          \uElseIf{($e_0 < e_r$)}{
          \eIf{($e_r < e_e$)}{$\mathbf{v_{n}}$ = $\mathbf{v_e}$\;}{$\mathbf{v_{n}}$ = $\mathbf{v_r}$\;}
          }
          
          \uElseIf{($e_{n} < e_r < e_{n-1}$)}{
          \eIf{($e_{oc} > e_r$)}{$\mathbf{v_{n}}$ = $\mathbf{v_{oc}}$\; }{$\forall \, i \in [1, n] \hspace{3mm} \mathbf{v_{i}} = s.\mathbf{v_i} $\;}}
          
          \uElseIf{($e_r > e_{n}$)}{
          \eIf{($e_{ic} > e_r$)}{$\mathbf{v_{n}}$ = $\mathbf{v_{ic}}$\;}{$\forall \, i \in [1, n] \hspace{3mm} \mathbf{v_{i}} = s.\mathbf{v_i} $\;}}
          sort the simplex\;
          \eIf{$\mathbf{v_{0new}} > \mathbf{v_0}$}{iter = 0}{iter = iter + 1}
          Update 'stop' from Algorithm \ref{algo:stop_criteria}
          }
         \Return $\mathbf{v_0}, e_0$
        \caption{Single start of the Nelder-Mead optimization algorithm}
        \label{algo:NM_single_op}
    \end{algorithm}
    
    \subsubsection{Pros and cons of the \ac{NM}-algorithm}
    \label{section:pro_con}
    The Nelder-Mead algorithm is quite straightforward to model the optimization problem for mechanism design. This allows us to design a general methodology for optimizing any parallel mechanism. As it is a derivative-free algorithm, we can introduce complex objective functions that are hard to formalize. An example is the quality index, \ac{gqi}, defined in Section \ref{subsection:constraints}. Also, as \ac{NM}-algorithm is a local search algorithm, it returns a stationary point in a considerably low time compared to the currently implemented global optimization methodologies. This makes it possible for the designer to structure an objective function that is computationally expensive. Also, the constraints can be constructed in modular way, allowing to experiment with different constraints at any stage of the development. Another important advantage of the Nelder-Mead algorithm relevant to the mechanism design is its geometric search method. The basis of optimization space in \ac{NM}-algorithm \changes{is} the optimization variables themselves. It is logical to use this method because the next best design parameters are chosen as a result of the combination of parameters of previous simplex, rather than using complex methods to represent a mechanism in the optimization space which may not have geometrical explanation for choosing the next best proposal (e.g: chromosomes in Genetic Algorithm). We can also tune the exploring parameters, i.e., the reflection, expansion, contraction and shrinkage coefficients, with human intuition and some prior knowledge about the importance of different parameters. \\
    
    \indent Though suitable for our application, there are certain disadvantages of using the NM algorithm, too. Under some hypotheses, the algorithm has proof of convergence up to dimension 2 \cite{lagarias_convergence_1998} and has no proof for convergence beyond 2-dimensional optimization. If not implemented correctly, it gets into a collapsing simplex patterns, thus converging to a non-stationary solution \cite{mckinnon_convergence_1998}. The convergence highly depends on the initial size of the simplex and the choice of the coefficients, as discussed in \cite{wang_parameter_2011}. Despite these shortcomings, the \ac{NM}-algorithm is useful in our case as the aim is not finding the absolute optimized design parameter but to satisfy all constraints and then get an acceptable quality of performance. Indeed, it has been implemented in various applications with great success \cite{luersen_globalized_2004, niegodajew_power_2020}. Different convergent variants have also been proposed to get around the premature convergence \cite{byatt_convergent_2000}, allowing the algorithm to explore extra points in case of near collapse. \\
    
    To get better results, local search of the NM algorithm is complemented with a multi-start technique for a global search in the optimization space, as discussed in the next section. 
    \subsection{Global search algorithm}
    The \ac{NM} algorithm combined with other global search methods such as  low-discrepancy points \cite{rudolph_proposal_2008}, genetic algorithm \cite{durand_combined_1999} and Powell optimization \cite{elleithy_globalization_2008} have been proposed in the past. We implement a multi-start Nelder Mead algorithm with low discrepancy points \cite{Nieder92, FanWan94, Opt10} for exploring a global optimization space. In this method, we execute the NM algorithm with different initial simplexes. It is very important to have a uniformly distributed initial simplexes over the optimization space, in order to explore the maximum area of the optimization space.
    
    \subsubsection{Initial simplexes for multi-start}
    \label{subsection:multi_start_NM}
    An easily implementable way to obtain a sampling set $\mathcal{O}_M \subset \mathcal{O}$ is {\em Monte Carlo sampling} with a uniform distribution (see, e.g., \cite{HamHan64}), i.e., {\em random sampling}. Unfortunately, it is known \cite{FanWan94} that the resulting points have the tendency to form clusters, particularly in high-dimensional contexts, which undermine the uniformity of the discretization. A better choice consists in having the $\, M $ points of the discretization $\, \mathcal{O}_M $ of $\, \mathcal{O} $ spread ``well-uniformly''. In particular, it is desirable that the points be close enough to one another, without leaving space regions under sampled. To this end, as done in \cite{Opt10,AlCervSangTNN07}, one can  use certain deterministic sampling techniques. The properties of such techniques are detailed in \cite{Opt10}. The work in \cite{Opt10} suggests that an efficient way of generating uniformly scattered deterministic sets of points consists in taking finite portions of so-called {\em low-discrepancy sequences} such as the {\em Halton sequence}, the {\em Hammersley sequence} and the {\em Sobol sequence}. The reported work utilizes initial simplexes chosen from the Sobol sequences, as they prove to be more uniformly distributed.\\
    
\begin{figure}[ht]
    \centering
    	\begin{subfigure}{0.48\textwidth}
    		\centering
    		\includegraphics[width=\textwidth]{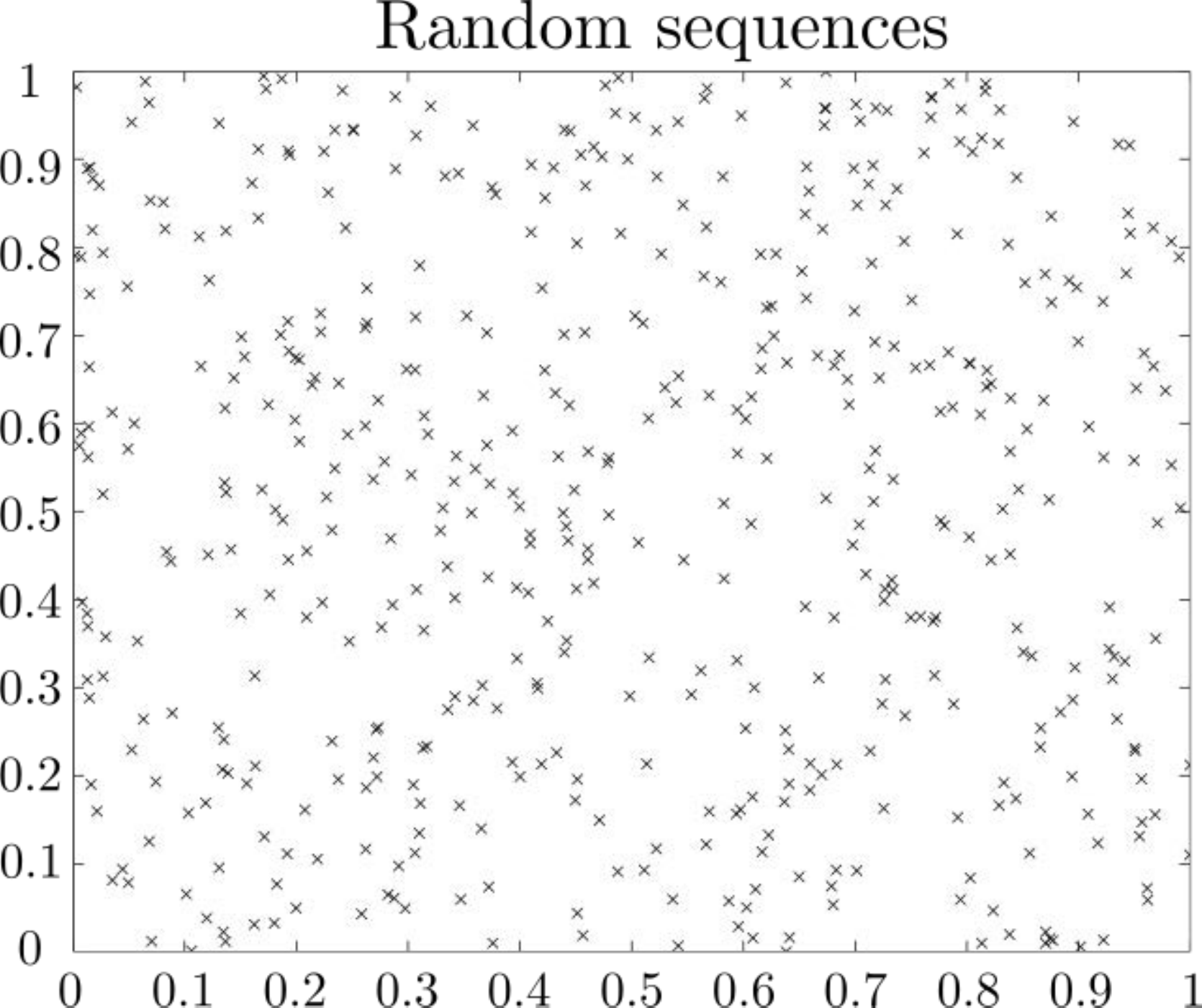}
    	\end{subfigure}
    	~
    	\begin{subfigure}{0.48\textwidth}
    		\centering
    		\includegraphics[width=\textwidth]{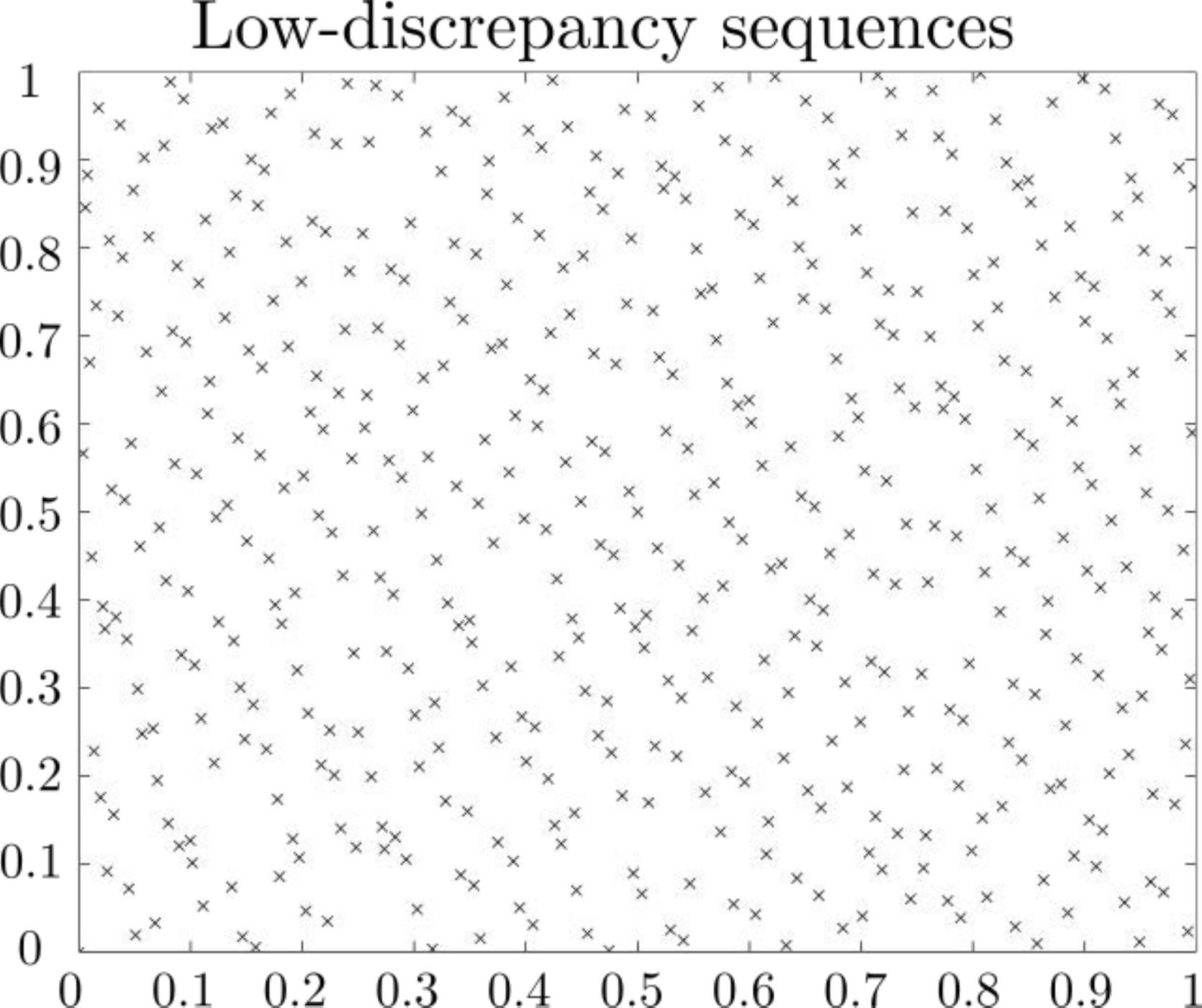}
    	\end{subfigure}
    	\caption{Comparison between random and low-discrepancy sampling of the unit square \cite{AlCervSangTNN07}}
    	\label{fig:sequences}
    \end{figure}

    Figure~\ref{fig:sequences} from \cite{AlCervSangTNN07} shows the comparison between a sampling of the 2-dimensional unit cube by a sequence of 500 points i.i.d. according to the uniform distribution and by a sampling of the same cube obtained via a low-discrepancy sequence (in this case,
    the {\em Sobol sequence} \cite{Sobol67}). It can be clearly seen how the space is better covered by the second sequence, as well as how the largest empty spaces among the points appear in the first sampling scheme.
    
    \subsection{Cascade optimization}
    \label{subsection:coarse_fine_search}
\begin{figure}[H]
    \centering
        \includegraphics[height = 4in]{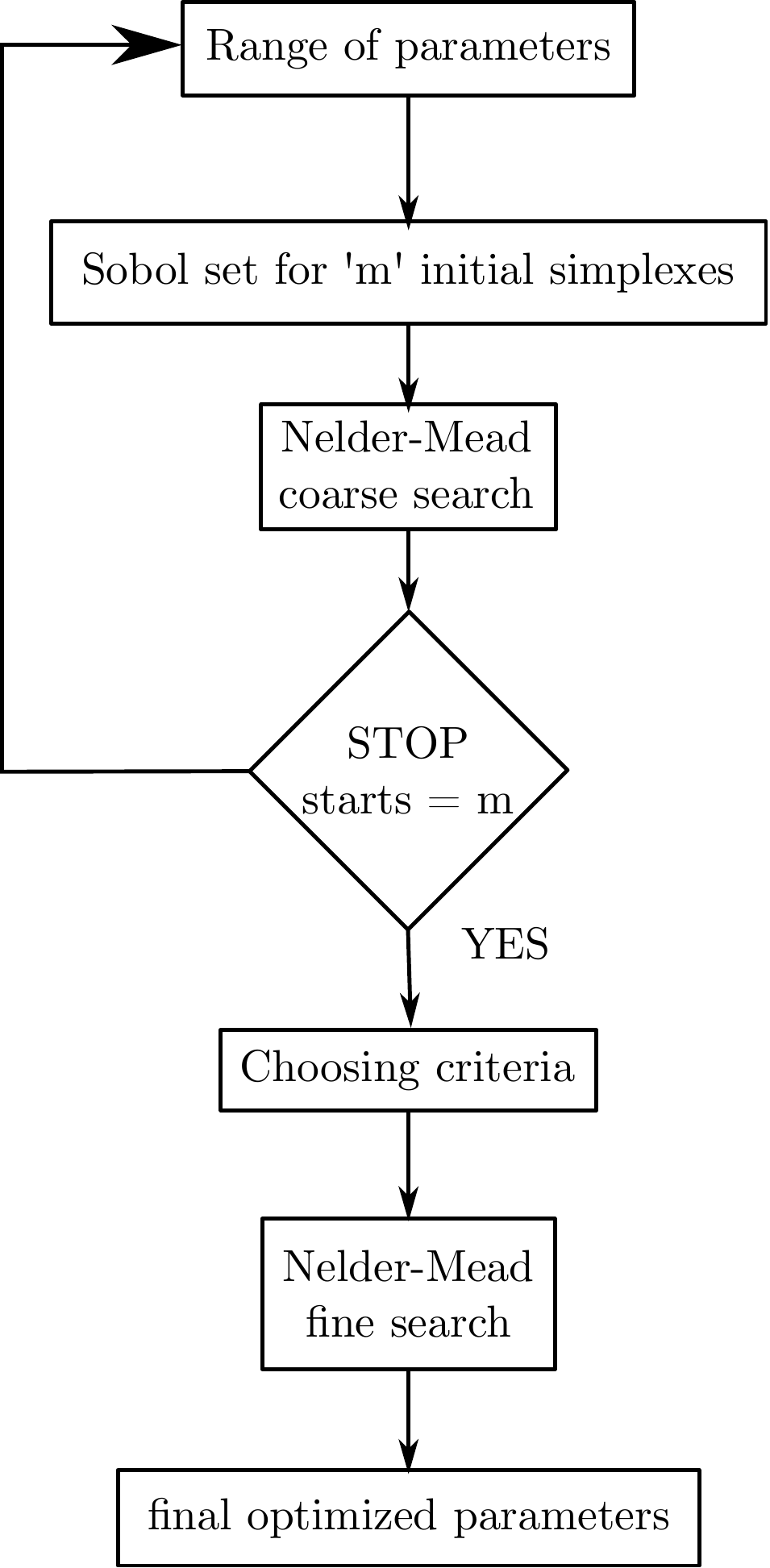}
        \caption{The flowchart for the complete implemented optimization methodology}
        \label{fig:flowchart_multiNM}
    \end{figure}

    In a normal execution of the Nelder-Mead algorithm, the iteration stops either when the simplex has shrunk to a desirable size with near the same evaluations or if we have encountered the same best point for preset allowable maximum iterations, refer to Algorithm \ref{algo:stop_criteria}. In an attempt to decrease the time for local convergence, allowing us to explore more initial simplexes, we adapt a methodology inspired from the practice of rough and fine turning in lathe machines. In general, when we want to remove the excess stock from the workpiece as rapidly as possible, we increase the feed rate and do not focus on the finish of the work. Later, when we are close to desired dimensions, the feed is decreased and now the focus is shifted on the finishing of the work. Figure \ref{fig:flowchart_multiNM} illustrates the complete flow of the algorithm. In the beginning, the simplexes taken from the Sobol sequence are initialized in multi-start NM-algorithm and a coarse search is performed for the local optima. Later, the local optima from some chosen initialized simplexes are used to implement stricter stopping criteria, allowing them to converge to a stationary point with finer quality. Fundamentally, we are discarding the local optima that do not promise a good evaluation even after a longer search, decreasing the computational time considerably. Also, as we already have an optimized vertex as an initial simplex, we can build the rest of the vertices as per our choice, thus controlling the size of the initial simplex.
    
    \subsubsection{Coarse search} 
    In the coarse search, we want to accelerate the local convergence, allowing us to maximize the number of starts in our optimization methodology. This is done by using a coarser search space and relaxing the stopping criteria. In the coarse search, the output space is discretized with an interval 10 times larger than in the finer search. This drastically brings down computational time. The aim of the coarse search is to find the simplexes that lie on relatively steeper slopes in the optimization space. By relaxing the stopping criteria, the maximum iterations allowed to repeat with the same evaluation is capped at 10 which helps in terminating the local search faster. One of such coarse implementation is detailed in Algorithm \ref{algo:coarse_fine}, where the condition of incrementing the iteration is changed. We implement a condition that the new evaluation found is better than the previous only if it exceeds the previous evaluation by 5\% and the algorithm stops as soon as we have 90\% of the maximum expected value.\\
    
    Unlike other optimization problems, in the PKM design the maximum optimized evaluation is known. For example, if we are discretizing the output space in 1000 points and are rewarding a value of 1 for a feasible point and 0 for infeasible points, then the maximum evaluation of such rewarding strategy is 1000. This fact is so useful that we can now have criteria related to the maximum expectation and the current evaluation. In the coarse search, we implement a constraint such that if an evaluation is greater than 80\% of the maximum evaluation, then terminate the iterations. This particular methodology lowers the optimization time drastically when the constraints are not too strict, and we have many parameters satisfying the constraints. It is interesting to note that this methodology can be used irrespective of the rewarding strategy. 
    
    \subsubsection{Fine search}
    In the fine search, we filter the different local optima obtained from the coarse search. The evaluations of the local optima are arranged in increasing order, and the top 10\% of the collected optima are chosen for further evaluation. In the fine search, we implement stricter stopping criteria, change the constraint of maximum expected evaluation to 100\% and discretize the output space with a 10 times finer interval. The margin that is considered as an improvement is lowered to 1\%. These changes directly affect the computational time and take much longer time with increasing dimension of the output space. All the optimized parameter sets from the NM-algorithm with finer constraints are compared, and the best point is proposed as an optimized parameter of the PKM.
    \vspace{0.1cm}

    \begin{algorithm}[H]
        \SetAlgoLined
        \KwResult{Optimised parameter set $\mathbf{v_0}$}
          input : Initial set of simplexes\; 
          $e_0$ : the best evaluation from the previous iteration\;
          $e_{max}$ : Maximum expected evaluation\;
          limit : The percentage of maximum evaluation that is considered best\;
          For coarse search\;
          max\_iter = 3\,n\;
          margin = 1.05 ... (suggesting $\geq$5\% increment is considered improvement)\; 
          limit = 0.8 ... (suggesting that 80\% of maximum evaluation is a criterion to stop)\;
          For fine search\;
          max\_iter = 10\,n\; 
          margin = 1.01 ... (suggesting $\geq$1\% increment is considered improvement)\;
          limit = 1\;
          stop = 0\;
          \While{stop = 0}{
          Perform algorithm \ref{algo:NM_single_op} except for last step of checking stop from algorithm \ref{algo:stop_criteria}\;
          Perform algorithm \ref{algo:evaluation} with finer intervals\;
          \eIf{$e_{new}$ $\geq$ margin$\times e_0$}{iter = 0}{iter = iter + 1}
          \If{iter $\geq$ max\_iter}{\Return stop = 1\;}
          \If{$e_{new} \geq$ limit$\times e_{max}$}{\Return  stop = 1\;}
          }
         \Return $\mathbf{v_{0}}$ from the algorithm \ref{algo:NM_single_op}
        \caption{Implementation of coarse and fine local search criteria}
        \label{algo:coarse_fine}
    \end{algorithm}
    \begin{algorithm}[H]
        \SetAlgoLined
        \KwResult{Optimized parameter set of the mechanism and its evaluation}
          Assuming we have `m' starts for a `n' dimensional optimization problem\;
          Choose m.(n+1) valid n-dimensional points from the Sobol set generated\;
          Choose `k' local optima for further fine search, generally, k $\leq$ 0.1\,m\;
          \For{start = 1:m}{
          Initial simplex = $\{\mathbf{v_{(m-1).(n+1)}} ... \mathbf{v_{mn + m -1}}\}$\;
          Implement Single start from Algorithm \ref{algo:NM_single_op} with coarse search from Algorithm \ref{algo:coarse_fine}\;
          $\mathbf{v}_{chosen}(start, 1:n+1) = [\mathbf{v}_0, e_0]$\;
          }
          sort $\mathbf{v}_{chosen}$ by evaluation of the corresponding parameter set\;
          \For{fine\_start = 1:k}{
          Generate n more parameter sets around $\mathbf{v}_{chosen}$(fine\_start)\;
          Implement Single start from Algorithm \ref{algo:NM_single_op} with fine search from Algorithm \ref{algo:coarse_fine}\;
          $\mathbf{v}_{fine}$(fine\_start, 1:n+1) = $[\mathbf{v}_0, e_0]$\; 
          }
          sort $\mathbf{v}_{fine}$ by evaluation of the corresponding parameter set\;
         \Return $\mathbf{v}_{fine}[1, 1:n]$, $\mathbf{v}_{fine}[n+1]$
        \caption{An example of implemented multi-start optimization}
        \label{algo:multi_start}
    \end{algorithm}

 \section{Results and discussion}
\label{chapter:implementation}
The optimization algorithm detailed in the work was used to optimize two different parallel mechanisms to validate the general implementation. The mechanisms chosen for optimization are widely used in the industry, and the relevance of the objective function chosen is also detailed in this section. An open source implementation of the proposed algorithm and the examples are available at: \url{https://github.com/salunkhedurgesh/ParaOpt}.

\subsection{1 dof lambda mechanism}
\label{subsection:lambda}
\begin{figure}[H]
    \centering
    \includegraphics[width = 0.4\textwidth]{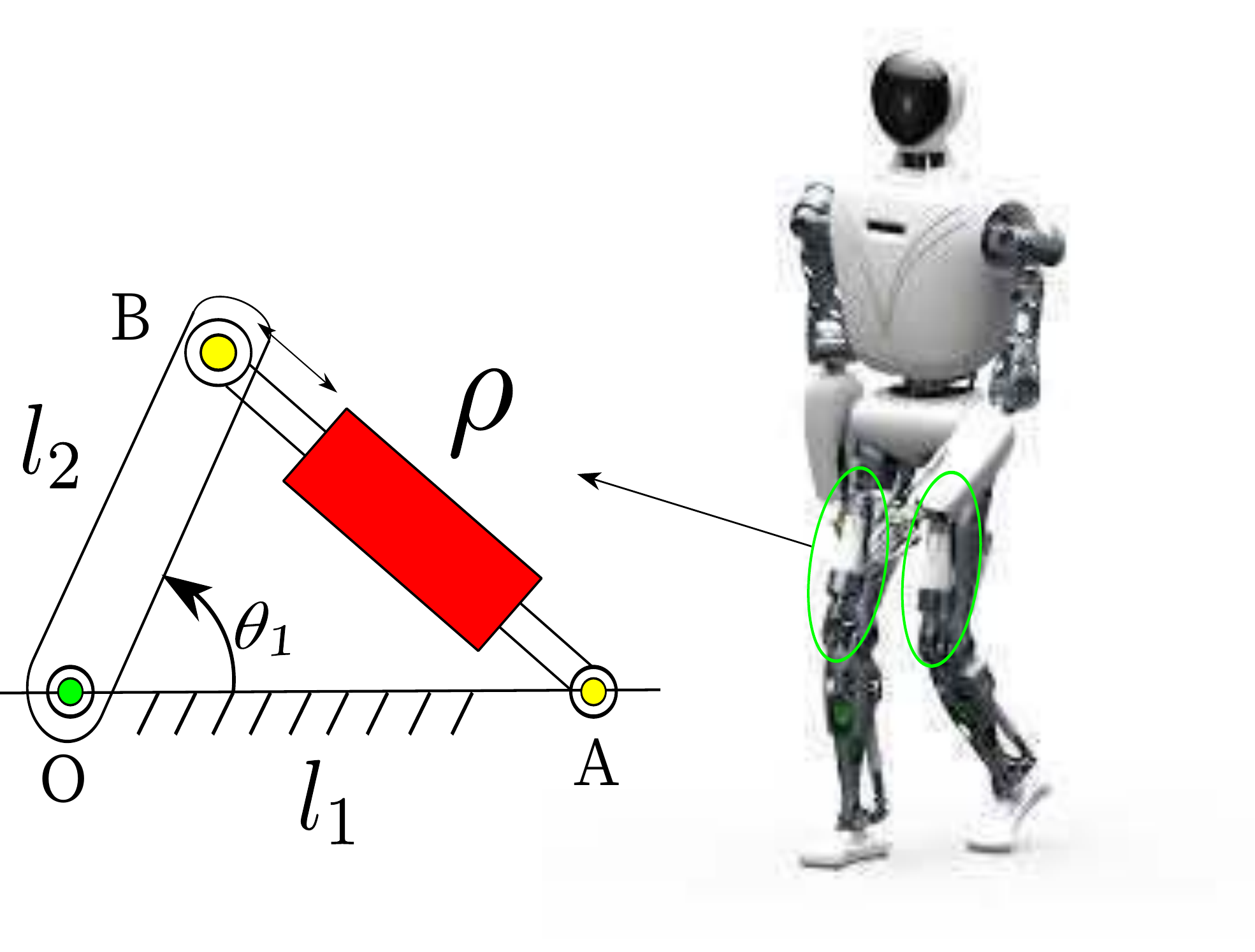}
    \caption{1-dof lambda mechanism with real life implementation}
    \label{fig:lambda_mech}
\end{figure}

The lambda mechanism is a single closed loop (1-RR\underline{P}R) mechanism and is used in the legged robots as an abstraction of revolute joint \cite{lohmeier_modular_2006, bartsch_development_2016, esser2021design} as shown in figure \ref{fig:lambda_mech}. This mechanism is used for a stiffer actuation where a compact, but powerful force is required, and non-linear transmission characteristics are desirable. The constraint equations are straightforward in this case and have been extensively discussed in \cite{kumar_modular_2019}. The mechanism was optimized by using the value of the determinant of the Jacobian matrix which is a scalar for the given case, $j$, as the GCI and a modified VAF. For the lengths and variables shown in figure \ref{fig:lambda_mech}, the calculations these measures are:
\changes{
\begin{equation*}
\begin{aligned}
    \rho^2 &= l_1^2 + l_2^2 - 2l_1l_2 \cos(\theta)\\
	j &= l_1 l_2 \dfrac{\sin(\theta)}{\rho}\\
	\text{GCI}_i &= j\\
	\text{VAF}_i &
    \begin{matrix}
    \begin{rcases}
    &= \dfrac{1}{1 + \sqrt{2}(\textit{j} - 1)^2} \\ &= 0\end{rcases} & \begin{matrix}
    \text{VAF}_{min} < j < \text{VAF}_{max} \\ \text{otherwise}
    \end{matrix}
    \end{matrix}\\
    \text{GCI} &=  \dfrac{\sum\limits_{i=1}^{n}\text{GCI}_i}{n}\\
	\text{VAF} &=  \dfrac{\sum\limits_{i=1}^{n}\text{VAF}_i}{n}
\end{aligned}
\end{equation*}
}
\begin{table}[H]
    \centering
    \resizebox{0.9\textwidth}{!}{
    \begin{tabular}{c|c|c|c}
        \hline Parameters & Value & Parameters & Value \\ \hline
         optimization dimension & 1 & Range of parameter & [1, 4]\\
         Number of starts & 100 & Number of iterations & 10\\
         Objective choice & Workspace, GCI, VAF & Velocity amplification range & [0.3, 3]\\
         Workspace ($\theta_1$ range) & $45^0$ to $135^0$ & stroke ratio & 1.5 \\ \hline
    \end{tabular}
    }
    \caption{The parameters set for the optimization of 1-dof lambda mechanism}
    \label{table:lambda_para}
\end{table}
\begin{figure}[H]
    \centering
    	\begin{subfigure}{0.46\textwidth}
    		\centering
    		\includegraphics[width = \textwidth]{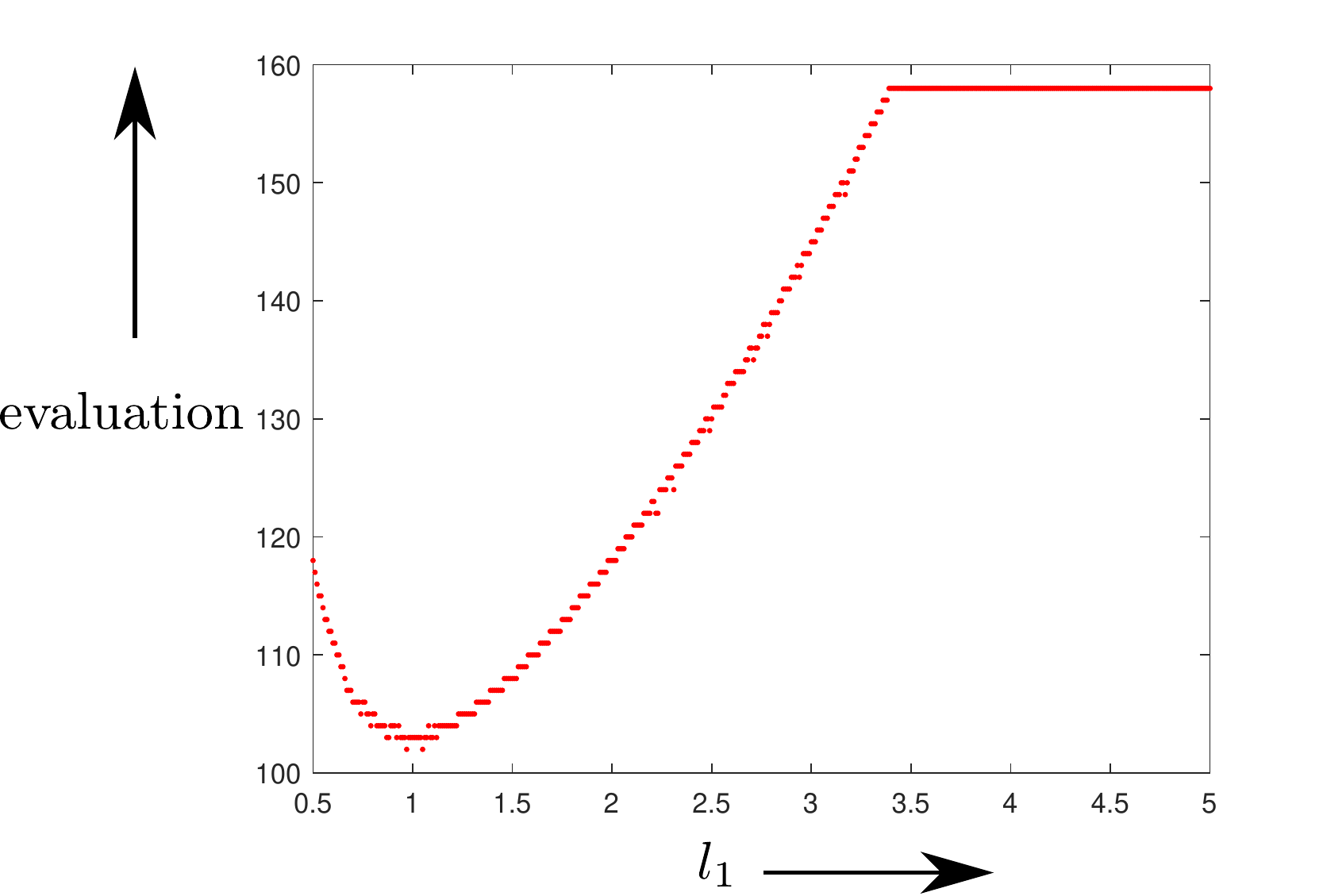}
            \caption{Evaluation of the objective function with varying parameter range(higher the better)}
            \label{fig:lambda_trend}
    	\end{subfigure}
    	~
    	\begin{subfigure}{0.46\textwidth}
    		\centering
    		\includegraphics[width = \textwidth]{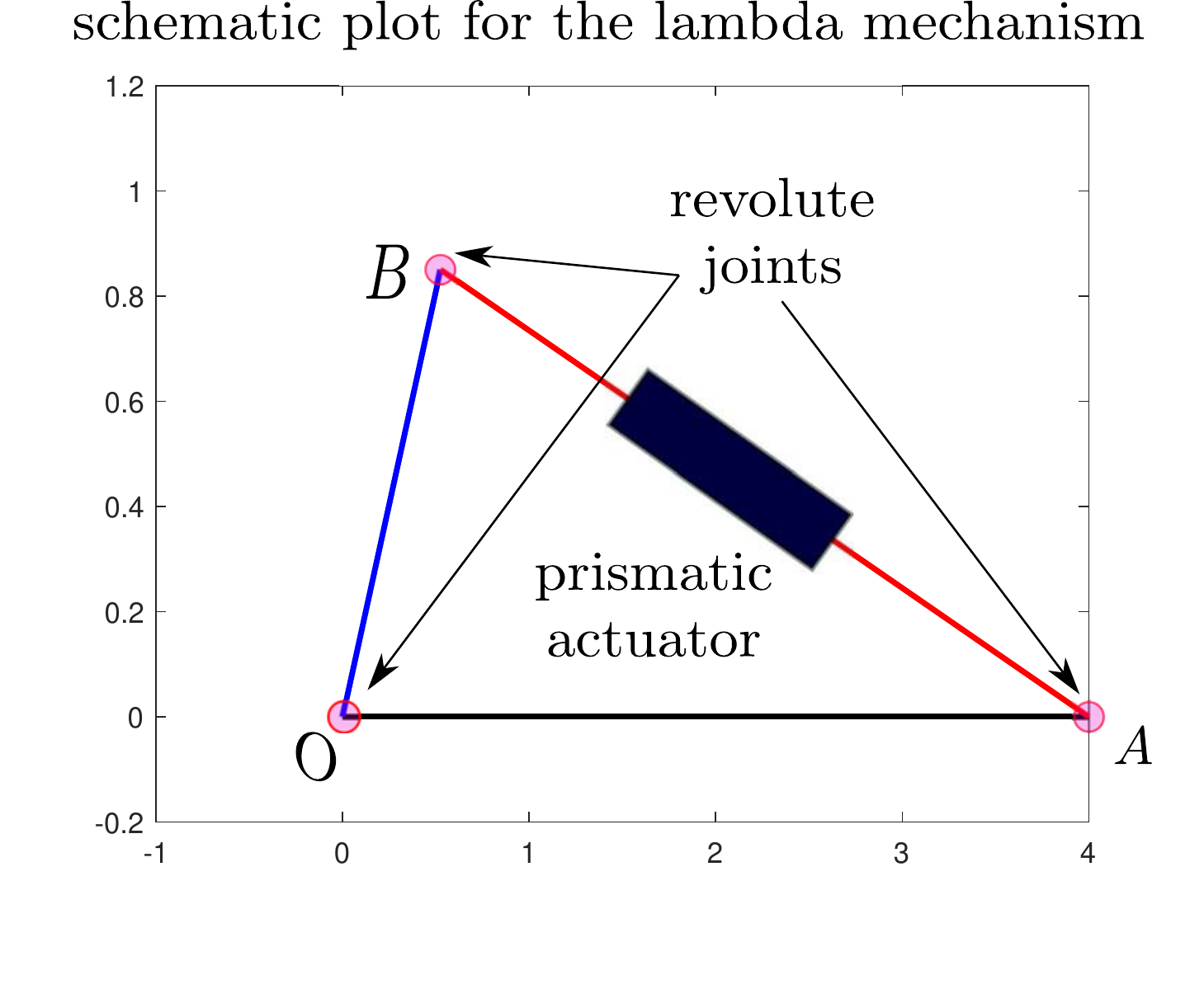}
            \caption{The schematic diagram of optimized lambda mechanism}
            \label{fig:lambda_result}
    	\end{subfigure}
    	\caption{Evaluation function plot and the schematics of the lambda mechanism for optimized length}
    	\label{fig:lamda_recent}
\end{figure}
In this mechanism, the length of $l_1$(OA) is optimized with respect to $l_2$(OB) and 3 different objective functions were used with parameters given in Table \ref{table:lambda_para}. In the first attempt, the workspace was maximized in order to find a good length such as to cover the revolute joint's travel from $45^0$ to $135^0$. Later, the \changes{GCI and VAF} were used as objective functions. The acceptable velocity amplification range for the mechanism was from 0.3 to 3. The stroke ratio, i.e. the ratio of length in full extension by length in no extension of the prismatic actuator, was $\frac{3}{2}$. 100 different single starts of local Nelder Mead optimization were used to tend towards a better global optimum, and the number of operations to be continued for the same evaluation in a single start were limited to 10 iterations. For all the objective functions, there are multiple solutions with equal evaluation. It was observed that $l_1$ = 4 was suggested as the global optimum while optimizing for all the different objective functions. As the optimization dimension was only 1, this operation was very fast and performed 100 coarse single starts and 10 refined starts in 21 seconds. Figure \ref{fig:lambda_trend} shows the plot for evaluation with different parameters. It can be observed that the evaluation increases till a certain value(=3.39) and then stays constant. This value is in fact the maximum possible evaluation in an ideal case, too. The results for the optimization of 1-dof lambda mechanism are summarized in Table~\ref{table:lambda_result}.

\begin{table}[H]
    \centering
    \resizebox{0.9\textwidth}{!}{
    \begin{tabular}{p{5cm}|p{5cm}|p{5.3cm}}
        \hline \hspace{1.5cm}Parameters & \changes{GCI} & \changes{VAF} \\ \hline
        Time for 1 coarse evaluation & 1 second & 1 second \\ 
        Time for single coarse start & 0.01 seconds & 0.01 seconds \\ 
        Time for one fine evaluation & 5 seconds & 3.1 seconds \\ 
        Time for single fine start & 0.04 seconds & 0.02 seconds \\ 
        Best point($l_2$) & 4 & 3.4 \\ 
        Best actuator range & [3.37 4.76] & [2.78, 4.17] \\\hline
    \end{tabular}
    }
    \caption{\changes{The results for the optimization of 1-dof lambda mechanism}}
    \label{table:lambda_result}
\end{table}
\subsection{2 dof RCM mechanism}
\label{subsection:rcm_implement}
\begin{figure}
    \centering
    \includegraphics[width = 0.5\textwidth, height = 3in]{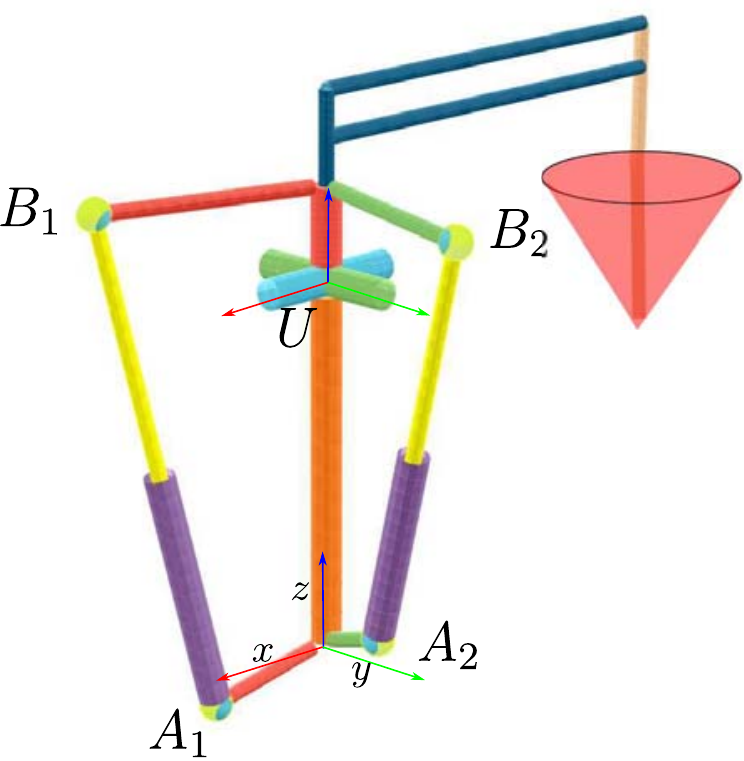}
    \caption{\changes{The parameters to be optimized in 2UPS-1U}}
    \label{fig:2UPSpara}
\end{figure}

To extend the optimization algorithm in its application, we decided to optimize a widely used 2-dof parallel mechanism, 2U\underline{P}S-1U. The mechanism can be assembled with double parallelogram to become a remote center of motion in order to facilitate tool mounting away from the actuators. This class of mechanisms have been used for medical applications \cite{michel_new_2020} as well as in implementing joint modules in humanoids (see~\cite{lenarcic_kinematic_2019, 8625046} for application as ankle joint and~\cite{esser2021design, 2022_Boukheddimi_rh5v2} for application as torso joint). This mechanism has three legs: the two first legs have 6 lower pair joints and include actuators. The third leg is a motion constraint generator, and the nature of the degree of freedom of the whole mechanism depends on the joint and its location. In the 2U\underline{P}S-1U, the universal joint in the third leg defines the two \changes{axes} of rotation as well as the center of rotation, which can be suitably displaced by using a parallelogram joint as shown in figure \ref{fig:2UPSpara}. The main advantage of such a manipulator is that it is free of constraint singularities and provides a rigid center of rotation. These advantages are very useful in surgeries and applications requiring precise motion in an intricate environment.
\changes{The first joints in leg 1 and leg 2 with respect to the base can be given as: 
\begin{align*}
    A_1 = \begin{bmatrix} a_1\cos\phi_1 \\ a_1\sin\phi_1 \\ h_1 \end{bmatrix}, \, A_2 = \begin{bmatrix} a_2\cos\phi_2 \\ a_2\sin\phi_2 \\ h_2 \end{bmatrix}
\end{align*}
where, $a_i$ is the distance of the first joint of $i^{th}$ leg from the origin of the base frame and $\phi_1$ is the angle between the xy-projection of vector from the origin of the base frame to the joint and the x-axis. Similarly, $\phi_2$ is the angle between the xy-projection of vector from the origin of the base frame to the joint and the y-axis. The joints of each leg are at height $h_1$ and $h_2$ respectively.
The universal joint (U) in the motion constraint generator leg is given as $[0, 0, t]^T$ with respect to the base frame.
The spherical joints in each leg are represented with respect to a frame with U as its origin and are given as:
\begin{align*}
    B_1 = \begin{bmatrix} b_1\cos\psi_1 \\ b_1\sin\psi_1 \\ h_3 + t \end{bmatrix}, \, B_2 = \begin{bmatrix} b_2\cos\psi_2 \\ b_2\sin\psi_2 \\ h_4 + t\end{bmatrix}
\end{align*}
where, $b_i$ and $\psi_i$ are used to express the spherical joints in the legs and have similar interpretation as that of $a_i$ and $\phi_i$. The joints of each leg are at height $h_3 + t$ and $h_4 + t$ respectively.}\\
Thus, the mechanism can be parameterized by 13 parameters after assuming that the motion constraint generator lies on the z-axis of the base. The 13 mechanism parameters to be optimized, as shown in figure \ref{fig:2UPSpara} \changes{and detailed above are}: $[a_1, \phi_1, h_1, b_1, \psi_1, h_2, a_2, \phi_2, h_3, b_2, \psi_2, h_4, t]$. The optimization parameters and the constraints along with their range are shown in Table \ref{table:2UPS_opttable}.
\begin{table}[H]
    \centering
    \resizebox{0.9\textwidth}{!}{
    \begin{tabular}{c|c|c|c}
        \hline Parameters & Value & Parameters & Value \\ \hline
         optimization dimension & 13 & Range of $a_i$ & [0.25, 1.5]\\
         Range of $b_i$ & [0.25, 2] & Range of $\phi_i$ and $\psi_i$ & [-1.745, 1.745]\\
         Range of $h_i$ & [-0.1, 0.1] & Range of t & [1, 4]\\
          Number of starts & 200 & Number of iterations & 10 and 20\\
          Objective choice & Workspace, GCI, VAF & Velocity amplification range & [0.3, 3]\\
         Range of $b_i$ & [0.25, 2] & Range of $\phi_i$ and $\psi_i$ & [-1.745, 1.745]\\
         Workspace (in roll and pitch) & circle of radius 1 & stroke ratio & 1.5 \\  
         limits on spherical joints & $\pm \pi/6 radians$ & Collision constraint & considered\\ \hline
    \end{tabular}
    }
    \caption{\changes{The parameters set for the optimization of 2-dof RCM mechanism}}
    \label{table:2UPS_opttable}
\end{table}

\begin{figure}[H]
    \centering
    \includegraphics[width = 0.9\textwidth]{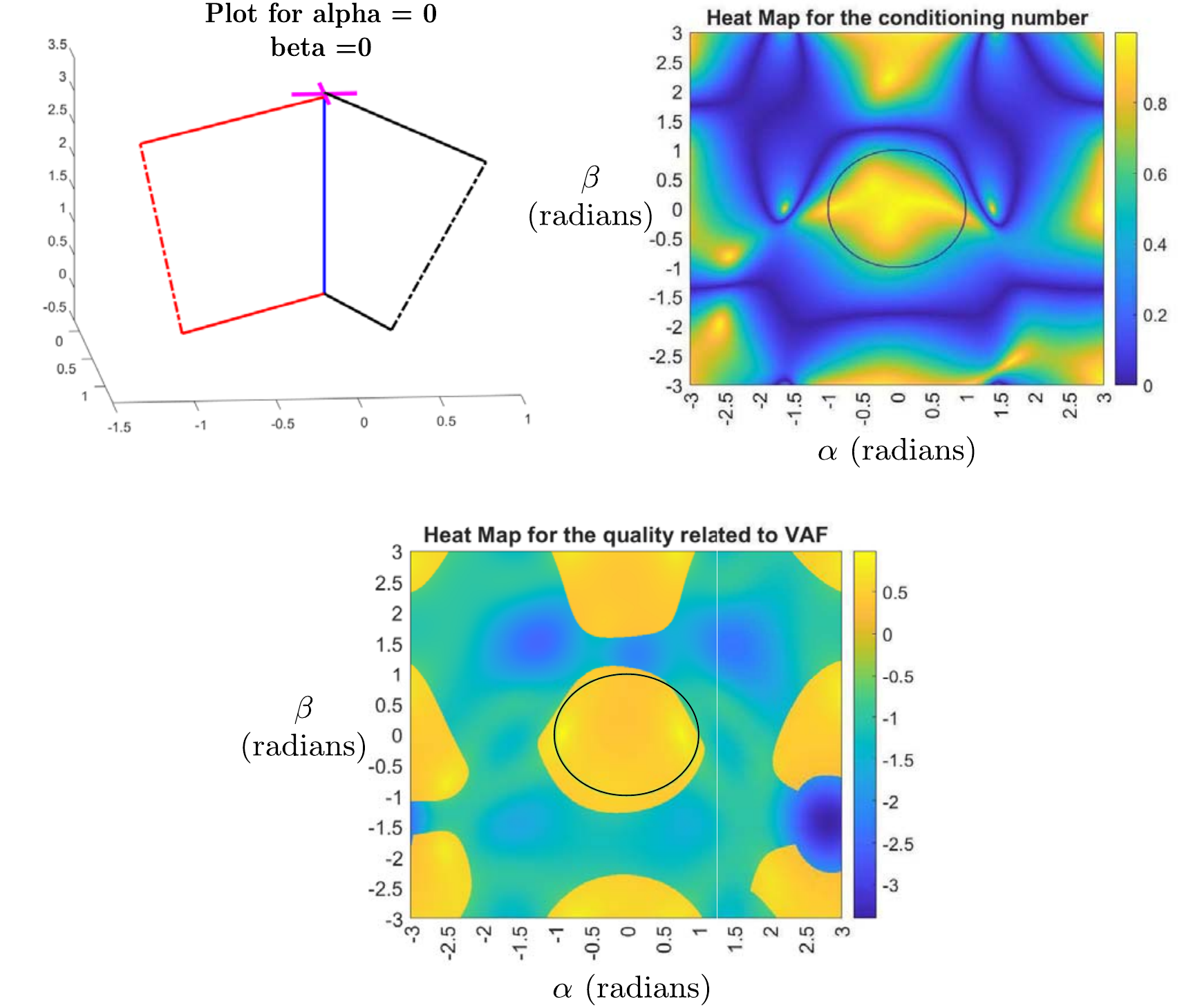}
    \caption{The schematic plot for the mechanism optimized for GCI and the heatmap for the evaluation. Calculation of GCI for this mechanism is discussed in \cite{chablat_workspace_2021}. The subfigure at the bottom is the heatmap for the VAF quality corresponding to the same parameters.}
    \label{fig:gci_opt}
\end{figure}

\begin{figure}[H]
    \centering
    \includegraphics[width = 0.9\textwidth]{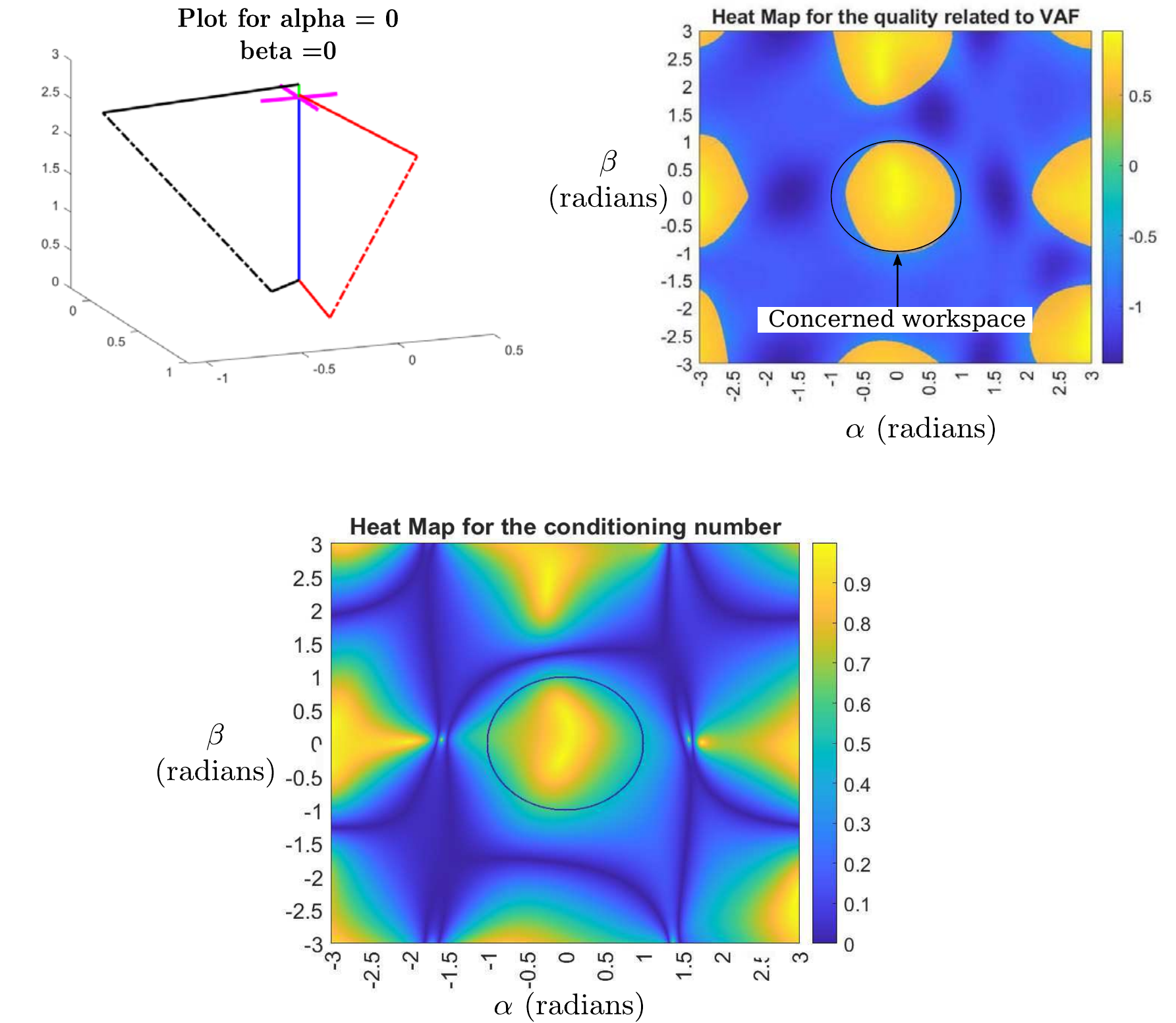}
    \caption{The schematic plot for the mechanism optimized for VAF and the heatmap for the evaluation. The subfigure at the bottom is the heatmap for the GCI corresponding to the same parameters.}
    \label{fig:vaf_opt}
\end{figure}

The optimization of this mechanism was much longer than the previous example because of the increase in the optimization space, number of dof and the workspace considered. The regular dextrous workspace for the given mechanism is discussed in details in \cite{chablat_workspace_2021}. The results vary depending upon the objective choice as well as the rewarding strategy. We present the results obtained (Table \ref{table:2UPS_opttable}) while optimizing for the GCI and rewarding a valid point in the workspace as 1 and 0 otherwise. The time required to evaluate one instance, i.e. one given set of parameters, was recorded along with the mean time for a single start, i.e. the complete operation till the algorithm stops to return the locally optimized parameters in \ref{table:2UPS_result}. It was further analyzed to note the impact of different objective choices on the total optimization time. It was also noted that the fine search took a very long time compared to coarse searches, thus emphasizing the efficiency of the algorithm. The results are presented in Table \ref{table:2UPS_result} and computational time is recorded on the same system and is to be used for comparison only. Figure \ref{fig:gci_opt} presents the schematic plot for the mechanism optimized for maximum \changes{GCI} along with the heatmap for the evaluation of GCI with the optimized parameters. Similarly, figure \ref{fig:vaf_opt} illustrates the schematic as well as the heatmap of the quality related to the \changes{VAF} for the corresponding optimized parameters. It is interesting to note from the schematics \changes{presented} in both figures that the optimized parameters tend towards an architecture such that the actuated legs are $\frac{\pi}{2}$ radians apart and align along the axes of the universal joint present in the motion constraint generator. This observation also suggests that we can use human intuition and experience to reduce the dimension of the optimization space, resulting in faster optimization and designs that are easy to manufacture.
\begin{table}[H]
    \centering
    \resizebox{0.9\textwidth}{!}{
    \begin{tabular}{p{5cm}|p{5cm}|p{5.3cm}}
        \hline \hspace{1.5cm}Parameters & GCI & VAF \\ \hline
        Time for 1 coarse evaluation & 14 seconds & 18.3 seconds \\ 
        Time for single coarse start & 291 seconds & 347.5 seconds \\ 
        Time for one fine evaluation & 50.5 seconds & 51 seconds \\ 
        Time for single fine start & 1072 seconds & 1077 seconds \\ 
        Best point & &\\$[a_1, \phi_1, h_1, b_1, \psi_1, h_2, a_2, \phi_2,$ $h_3, b_2, \psi_2, h_4, t]$ (refer figure \ref{fig:2UPSpara})& [1.13, -1.02, -0.06, 1.47, -1.01, -0.05, 0.72, 0.44, -0.02, 1.52, 0.54, 0.02, 3.04] & [0.68, -0.25, 0.08, 1.03, 0.1, 0.04, 0.25, -1, 0.01, 1.1, -1.45, 0.17, 2.4] \\ 
        Best actuator range & [2.54, 3.8] & [2, 3] \\
        evaluation \newline mean \newline standard deviation & GCI \newline 0.79 \newline 0.18 & VAF \newline 0.48 \newline 0.29\\
        maximum evaluation \newline configuration ($[\alpha, \beta])$& 1 \newline [0.39, 0.13]& 0.99 \newline [0, 0.43]\\
        minimum evaluation \newline configuration ($[\alpha, \beta])$& 0.318 \newline [0.86, 0.51]& -1.2 \newline [-0.99, 0.14]\\
        \hline
    \end{tabular}
    }
    \caption{\changes{The results for the optimization of 2-dof RCM mechanism}}
    \label{table:2UPS_result}
\end{table}


 \section{Conclusions}
    \label{chapter:conclusions}  
    In this paper, we present a novel optimization algorithm for parallel manipulators that is able to implement the joint limits and the collision of prismatic joints as constraints. The optimization methodology is also able to optimize the length of the actuator stroke, which enables the designer greater flexibility and clarity in the choice of the actuators. The algorithm uses geometrical traversing for optimization, which is very relevant for mechanism optimization. The algorithm implements a two-step search by combining a faster local search Nelder-Mead algorithm with initial simplexes spread over all the parameter space and then uses a finer search by using the locally optimized points in the step 1. The algorithm is general and can adapt to any non-redundant parallel mechanisms with prismatic as well as revolute joint. The paper presents two different mechanism optimization as an example to present the flexibility of the algorithm. The algorithm can be used in the systems that can propose different models based upon the requirements. In future works, the algorithm will be extended to multi-objective optimization to select the best architecture and corresponding design parameters of a robot for a given task.

\section*{Acknowledgment}
    The project received financial support from the NExT (Nantes Excellence Trajectory for Health and Engineering) Initiative and the Human Factors for Medical Technologies (FAME) research cluster. The third author acknowledges the support of M-RoCK project (FKZ 01IW21002) funded by the German Aerospace Center (DLR) with federal funds from the Federal Ministry of Education and Research (BMBF) respectively.

\bibliographystyle{elsarticle-num}
\bibliography{bibliography/references}
\end{document}